\documentclass{applemlr}

\usepackage{amsmath}
\usepackage{enumerate}
\usepackage{algorithm}
\usepackage{algpseudocode}
\usepackage{amsfonts}
\usepackage{amsthm}
\usepackage{cleveref}
\usepackage{diagbox}
\usepackage{colortbl}
\usepackage{amssymb}
\usepackage{xspace}
\usepackage{wrapfig}
\usepackage{adjustbox}
\usepackage{tabularx}
\usepackage{booktabs}
\usepackage{mathtools}
\usepackage{tikz}
\usepackage{enumitem}
\usepackage{silence}
\usepackage{dsfont}
\usepackage[table]{xcolor}
\usepackage[dvipsnames]{xcolor}
\usepackage{multirow}
\usepackage{makecell}
\usepackage{xfakebold}

\usepackage{amsmath,amsfonts,bm}

\def\eqref#1{equation~\ref{#1}}

\def\1{\bm{1}}

\DeclareMathAlphabet{\mathsfit}{\encodingdefault}{\sfdefault}{m}{sl}
\SetMathAlphabet{\mathsfit}{bold}{\encodingdefault}{\sfdefault}{bx}{n}

\definecolor{textgray}{HTML}{6E6E73}
\usetikzlibrary{positioning, calc}
\usetikzlibrary{decorations.pathmorphing}

\makeatletter
\patchcmd{\wrong@fontshape}{\@gobbletwo}{}{}{}
\makeatother
\numberwithin{equation}{section}
\makeatletter
\AtBeginDocument{
  \urlstyle{sf}
  
}
\makeatother

\definecolor{light}{RGB}{125, 125, 125}
\crefname{tcb@cnt@pbox}{code}{code}
\Crefname{tcb@cnt@pbox}{Code}{Code}
\crefname{assumption}{assumption}{assumption}
\Crefname{assumption}{Assumption}{Assumptions}

\newtcolorbox[auto counter]{pbox}[2][]{
  colback=white,
  title=Code~\thetcbcounter: #2,
  #1,fonttitle=\sffamily,
  fontupper=\sffamily,
  arc=2pt,
  colframe=bgcolor,
  coltitle=fgcolor,
  colbacktitle=bgcolor,
  toptitle=0.25cm,
  bottomtitle=0.125cm
}

\makeatletter
\newcommand\applefootnote[1]{%
  \begingroup
  \renewcommand\thefootnote{}%
  \renewcommand\@makefntext[1]{\noindent##1}%
  \footnote{#1}%
  \addtocounter{footnote}{-1}%
  \endgroup
}
\makeatother

\definecolor{cverbbg}{gray}{0.90}

\usepackage{url}
\usepackage{bbm}
\usepackage[textwidth=18mm]{todonotes}

\newcommand{\method}{SynthAPI\xspace}
\newcommand{\sapiSmallSub}{SynthAPI-SubOpt\xspace}
\newcommand{\sapiSmallOpt}{SynthAPI-Opt\xspace}
\newcommand{\sapiFull}{SynthAPI-Big\xspace}

\newcommand{\sapihold}{SynthAPI-Holdout\xspace}

\newcommand{\asn}[1]{}
\newcommand{\as}[1]{}
\newcommand{\mkn}[1]{}
\newcommand{\ravi}[1]{}
\newcommand{\cy}[1]{}

\usepackage[most]{tcolorbox}

\newtcblisting[auto counter]{promptbox}[2][]{
  colback=gray!5,
  colframe=black,
  title={Prompt~\thetcbcounter: #2},
  breakable,
  enhanced,
  listing only,
  listing options={
    basicstyle=\ttfamily\small,
    breaklines=true,
    breakindent=0pt,
    columns=fullflexible,
    keepspaces=true,
    upquote=true
  },
  #1
}
\crefname{tcb@cnt@promptbox}{prompt}{prompts}
\Crefname{tcb@cnt@promptbox}{Prompt}{Prompts}
\crefname{equation}{Equation}{Equations}
\Crefname{equation}{Equation}{Equations}

\usepackage{titletoc}
\titlecontents{section}
  [2.2em]                                   %
  {\bfseries}                             %
  {\contentslabel{2em}}                   %
  {}                                      %
  {\hfill\contentspage}                   %
  [\vspace{0.75ex}]                       %

\title{Scaling Laws for Tool-Use Agents via Controllable Environment Synthesis}

\author{Andrew Szot}
\author{Samira Abnar}
\author{Omar Attia}
\author{Abbas Kazerouni}
\author{Michael Kirchhof}
\author{Cheng-Yu Hsieh}
\author{Raviteja Vemulapalli}
\author{Alexander Toshev}

\affiliation{Apple}

\abstract{We explore how to build performant large language model (LLM) tool-use agents that complete complex, multi-step tasks. Specifically, we study how data composition in synthetically generated tool-use environments affects the training of such agents. To do so, we propose the SynthAPI pipeline, which explicitly controls the diversity of generated environments along the axes of tools, tasks, and the databases on which the tools operate. By systematically varying SynthAPI's generation controls in 458,752 tasks across 849 trained models of four different sizes, we derive a scaling law that predicts downstream tool-use agent generalization based on the scale and composition of the synthesized environments. These predictable relationships reveal several lessons for scaling tool-use agent data. As data budgets grow, scaling tool and database diversity becomes more effective than increasing the number of tasks, and increasing data complexity yields larger gains. Scaling according to these findings yields agents that outperform models trained on data from prior work in large-scale tool-use agent and synthetic environment generation across four evaluation benchmarks.
}

\metadata[Correspondence]{\sffamily Andrew Szot: \url{a_szot@apple.com}}
\date{\sffamily\today}

\begin{document}

\maketitle

\applefootnote{ \textcolor{textgray}{\sffamily Apple and the Apple logo are trademarks of Apple Inc., registered in the U.S. and other countries and regions.}}

\section{Introduction}
\label{sec:introduction}
Large language models (LLMs) have demonstrated impressive capabilities as tool-use agents that interact with complex, stateful environments \citep{trivedi2024appworld,patil2025bfcl}.
Such agents must select and compose tools over multiple steps while tracking state changes and generalizing to new environments with new APIs.
Progress in developing such agents depends on scaling high-quality, diverse training data.
Yet producing this data is challenging because multi-step tool interactions are scarce in internet text and require realistic tasks, diverse tools, and rich environment data.

Prior work addresses this challenge by using LLMs to synthesize tool-use data.
While many existing works focus on synthesizing tool-use data to train stateless, read-only tool-use agents \citep{schick2023toolformer}, work on scaling supervision for stateful tool-use agents is more limited.
Some approaches use LLMs to synthesize entire tool-use trajectories, including simulating the tool responses \citep{liu2024toolace,tang2023toolalpaca}.
These techniques have demonstrated impressive scaling \citep{nvidia2025nemotronagentic,lee2026environment}, but complex tool outputs and state transitions can be difficult to simulate reliably.
Other approaches synthesize code-backed executable tool environments \citep{wang2026agentworldmodel,song2026envscaler,xu2026envfactory,dong2026agent}.
However, these efforts primarily emphasize increasing data quantity, leaving open which properties of the generated data drive downstream performance.
Tool-use agent data varies in the diversity of available tools, the tasks they support, and the database states over which they operate.
Understanding how these properties affect generalization is necessary to determine what data to generate as training scales.

\begin{figure*}[t!]
\centering
\includegraphics[width=\linewidth]{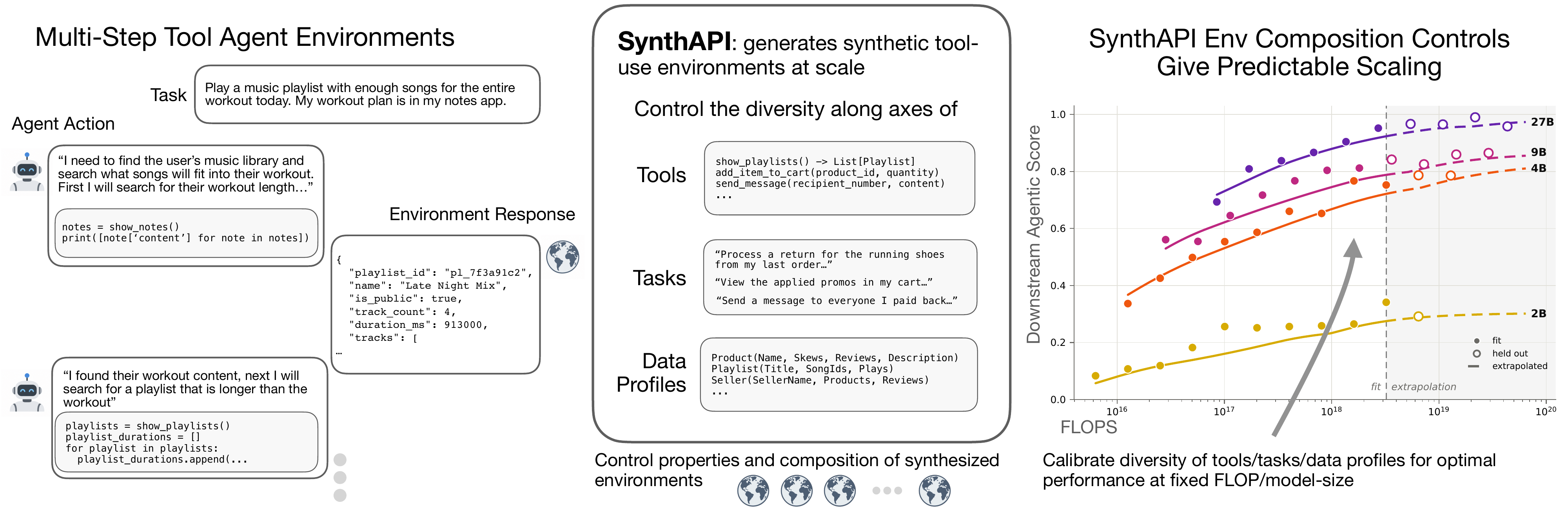}
\caption{
    Overview of \method. 
    On the left, we show the tool-use agent tasks we focus on, where given an instruction, an agent must reason and execute tool calls to complete the task. 
    In the middle, the \method pipeline generates synthetic tool-use agent environments at scale. 
    \method generates this data while controlling the diversity of the tools, tasks, and starting database states. 
    These controls allow deriving the scaling laws on the right (presented in detail in \Cref{fig:model-size-flops}).
}
\label{fig:appgen-overview}
\end{figure*}

To study these relationships, we introduce \method, a controllable synthesis pipeline for multi-step tool-use environments.
\method uses LLMs to implement apps comprising Python tools that read and modify structured databases. 
\method then generates tasks with initial database states, which we call data profiles and executable success verifiers.
The key distinction from prior executable environment synthesis approaches is that \method provides explicit controls over the diversity of tools, tasks, and data profiles.
These controls allow us to vary the composition of the synthesized training tasks and study how a fixed budget should be allocated among new tools, new tasks, and new data profiles.
\Cref{fig:appgen-overview} summarizes the pipeline and its controls.

We use this controllability to derive a predictive scaling law that uses data composition factors to predict downstream tool-use agent generalization.
We find that the data composition that best supports downstream generalization changes with scale.
Comparisons at fixed task budgets show that, as these budgets grow, expanding tool diversity becomes more valuable than generating additional tasks within the same set of tools.
Furthermore, repeating the same tasks with different data profiles also becomes more important than generating new tasks.
Training with more complex data profiles also yields larger gains at higher budgets.
These findings emphasize the importance of tool and data profile diversity over simply increasing the number of tasks.
We propose an environment composition dependent scaling law that captures these effects.
Across four model sizes, we hold out the runs in the top 25\% of training FLOPs and show that the law extrapolates to them, including runs with up to $16\times$ the largest FLOP budget used for fitting, achieving a mean absolute error of $0.0285$ on the normalized downstream score, compared with $0.1706$ for a scaling model that relies on the number of tasks alone and ignores the composition of environment properties.

With the optimal environment composition recipe determined by our proposed method, we show \method transfers to broad tool-use agent benchmarks across AppWorld, BFCL, and ASTRA-Bench \citep{trivedi2024appworld,patil2025bfcl,xiu2026astrabench}.
At a fixed data budget, the recipe identified by our analysis outperforms a suboptimal recipe that simply relies on scaling the number of tasks alone across all model sizes and benchmarks.
We find that agents trained on the \method data outperform agents trained on prior large-scale tool-use agent data \citep{nvidia2025nemotronagentic} by $30.5$ percentage points on average across the three benchmarks.
We also compare \method to strong per-benchmark data synthesis baselines, outperforming the strongest baseline at each model size by $7.6$--$11.6$ percentage points on AppWorld Test-Challenge and $2.0$--$3.7$ points on BFCL \citep{lee2026environment,xu2026envfactory,song2026envscaler}.
These results demonstrate the importance of properly scaling the data composition of synthesized environments through our scaling law enabled by the controllable environment generation in \method.

\section{Related Work}
\label{sec:related-work}

Synthetic supervision has been widely used to train LLMs for tool use.
Early methods train on demonstrations over predefined API collections \citep{schick2023toolformer,qin2023toolllm,chen2024agentflan}, while subsequent work generates supervision through live APIs and executable functions \citep{liu2024apigen,prabhakar2025apigenmt}.
Other approaches simulate API responses with LLMs \citep{liu2024toolace,tang2023toolalpaca} or combine data from multiple sources and synthesis pipelines \citep{zhang2025xlam}.
DIVE \citep{chen2026dive} analyzes tool diversity in evidence-based question answering, whereas \method studies multi-step read-write workflows over structured databases, a setting emphasized by recent tool-use agent benchmarks \citep{trivedi2024appworld,patil2025bfcl}.
Nemotron-Agentic \citep{nvidia2025nemotronagentic} scales LLM-simulated tool interactions to 335k trajectories.
SimSAT \citep{lee2026environment} extends such simulation to long-horizon tasks.
However, these simulators rely on an LLM to generate every tool response, making increasingly complex outputs and state transitions difficult to simulate reliably.
Thus, unlike these works, \method scales tool data through database-backed executable environments.

Like \method, recent work also synthesizes executable tool-use environments backed by persistent databases \citep{wang2026agentworldmodel,song2026envscaler,xu2026envfactory,dong2026agent}.
Unlike these works, \method couples this synthesis with explicit controls over the diversity and complexity of tools, task instructions, and initial database states.
These controls let us disentangle how training-data composition affects downstream generalization.
Our analysis goes beyond prior scaling studies that primarily vary the number of synthesized environments or examine domain diversity at a fixed task count \citep{tu2026scaleenv}.
Specifically, we study how to allocate a fixed data budget among new tools, additional tasks for existing tools, and alternative database states for existing tasks.
We further derive a joint scaling law that extrapolates aggregate downstream generalization to larger compute budgets and predicts which data allocations maximize performance.
We use these predictions to design a training recipe that improves agent performance over prior synthetic-data generation approaches.

Scaling laws characterize predictable relationships between model quality, model size, data volume, and training compute.
Early work established power-law behavior in language-model pre-training, while compute-optimal analyses studied how a fixed budget should be divided between parameters and tokens \citep{kaplan2020scalinglaws,hoffmann2022training}.
Related studies extend scaling analysis to transfer learning, fine-tuning, and reinforcement-learning post-training \citep{hernandez2021scalingtransfer,tan2025rlscaling}.
For tool-use models, prior work develops scaling laws for pre-training with agent data, using benchmark loss as the prediction target \citep{zhuang2025hephaestus}.
In contrast, \method predicts downstream task success from environment-generation factors, relating generalization to the scale, and diversity of generated tools, tasks, and data profiles.

\section{\method: Synthetic Environments for Tool-Use Agents}
\label{sec:appgen}

\method is a controllable pipeline for synthesizing interactive tool-use agent environments.
\method consists of two stages where first apps consisting of APIs and databases are coded in Python and next tasks are constructed using the functionality in these apps.

\subsection{API Generation}
\label{sec-api-gen} 

In the first phase of creating synthetic tool-use agent tasks, \method generates a wide diversity of tools to base tasks on.
\method does this by synthesizing ``apps'', which are a coherent collection of interacting tools that can be composed to solve realistic tasks.
This organization naturally produces dependencies among tools and mirrors deployment settings in which agents use tools to accomplish user goals. 
These apps are each independently implemented in Python code from scratch by an LLM, which we term the ``builder LLM''.
We use Gemini-3.1 Pro \citep{googledeepmind2026gemini31pro} with internet access disabled as the builder LLM.
We manually define a broad app taxonomy consisting of shopping, music, productivity, finance, fitness, messaging, utilities, and social.
The builder LLM starts to generate an app by generating ``metadata'' consisting of the app name, a short app description, the taxonomy category the app belongs to, and a list of functionalities.
The functionalities are short descriptions of interactions possible in the app, such as ``Initiate a return'' or ``Apply a promo code to your shopping cart'' in the case of a shopping app.

Next, \method generates a database schema for the app from the metadata to ground app functionality in complex stateful interactions.
For example, in a shopping app there need to be products, inventory, users, addresses, payment methods, reviews, carts, and orders for the tools in the app to implement meaningful behaviors.
Given the app metadata, the builder LLM generates a \texttt{db.py} file that contains Python \texttt{Enum} and \texttt{dataclass} definitions defining all data types necessary to instantiate the provided app functionalities.
These classes only specify data formats rather than instantiated database states.
The final step of app creation is generating an API layer that operates on the database schema.
From the contents of \texttt{db.py} and the app metadata, the builder LLM generates an \texttt{apis.py} file for the app.
The API file consists of a set of Python functions defining the possible operations in the app.
Each function takes as input a database object that tracks instantiated data schema objects, along with any other input arguments, which may consist of the defined data types.
These functions read or modify instantiated database objects and collectively define the agent's available tools.
In a shopping app, example resulting APIs may search products, inspect reviewers, or create an order.
Public functions from the API set are exposed to the agent as callable tools.

In summary, in the first phase, \method generates a set of apps, each of which contains app metadata, a data schema Python file, and an API definition file. All components are generated from scratch by the builder LLM. 
See \Cref{supp-app-gen} for all details and prompts on app generation.

\subsection{Task Generation}
\label{sec-control} 

The apps specify which tools and data schemas are possible for the agent to interact with, but they do not specify actual tasks to complete.
The second phase of \method therefore synthesizes actionable tasks for agents to complete using the provided APIs. 
A task is defined as a tuple $T = \bigl(x, v, \text{Db}^{0}, \pi^{*}\bigr)$, where $x$ is the task instruction, $\text{Db}^{0}$ is the starting database state, $v$ is verifier code that determines whether the task has been completed, and $\pi^{*}$ is an oracle solver program which solves the task from the starting state.
Next, we detail these components and how they are generated.

First, \method generates the task instruction by sampling an app and subsampling $F$ functionalities from its metadata functionality list.
The builder LLM is then prompted to generate $n$ task instructions that utilize these functionalities.
To increase instruction diversity, previously generated instructions are included as context for subsequent generations.
The instructions contain no ambiguity about the goal and require no additional information or interaction with an external user.

Next, given a task instruction $x$, \method generates the corresponding task verifier $v$.
The verifier is a Python function that takes as input the starting and current database state along with an optional answer string provided by the agent and outputs a binary indicator of whether the task instruction has been successfully completed in the current database state.
The verifier can also access historical information from the database state based on the state changes in the episode.
The builder LLM generates the verifier from the instruction and all the app data class and API definitions.

Provided $(x, v)$ and the app's \texttt{db.py}, \method then prompts the builder LLM to generate the initial database state $\text{Db}^{0}$ which consists of instantiated data objects from the app schema.
A first pass populates the data objects with the minimum data setup $\text{Db}_{\text{core}}^{0}$ required to make the task feasible.
For example, for the task ``buy a red mug", a red mug product and user account details need to exist in the shopping app for the task to be feasible.
Next, the data generator creates a batch of extra distractor data $\text{Db}_{1}^{0}$ to make the task more complex and realistic. 
$D-1$ such extra batches of data are created, thus creating $D$ total batches of data $\text{Db}^{0} = \text{Db}_{\text{core}}^{0} \cup \text{Db}_{1}^{0} \cup \dots \cup \text{Db}_{D-1}^{0}$.
From $(x, v, \text{Db}^{0})$, the builder LLM generates an oracle solver $\pi^{*}$ to check that the task is feasible to complete.
The oracle solver is a Python code block that executes the same APIs available to the agent to complete the instruction.
$\pi^{*}$ is generated using the entire database object $\text{Db}^{0}$, and thus it can directly access privileged identifiers and information, unlike the agent, which needs to query the database and tools to discover the right information.
Finally, \method runs a filtering process and rejects tasks that do not satisfy $v(\text{Db}^{0})=0$ and $v\!\left(\operatorname{Exec}(\pi^{*}, \text{Db}^{0})\right) = 1$.
See \Cref{supp-task-gen} for all details and prompts on task generation.

Each task $T_i$ is implemented in the standard Gymnasium environment format \citep{towers2024gymnasium}. 
The agent interacts with the environment by sending Python code blocks. This allows the agent to call multiple tools in one turn and to use loops, conditionals, and other programmatic operations over tool outputs.
Tool results are returned to the agent in JSON format, as is standard in other benchmarks.
\method exposes a wide variety of tools, and the tools required for a given task are not known in advance.
Rather than placing every tool definition in the initial context, we give the agent a small set of meta-tools for discovering and using the available tool interface.
\texttt{list\_apps} returns the available apps, \texttt{list\_tools} returns all tools within a selected app, and \texttt{get\_tool\_documentation} returns a tool's description, input and output arguments, types, and default values.
The list functions also return the apps and tools not necessary to solving the task, requiring the agent to ignore distractors.
Finally, the agent must call \texttt{complete\_task} when it thinks the task is finished, along with an optional answer string if the task requires an answer.
This discovery-based interface lets the agent load only the tools relevant to the current task, reducing context length while allowing it to transfer its learned behaviors to new tool sets across downstream benchmarks.
All tasks involve only tools in a single app and we defer generating multi-app tasks to future work.
However, we show agents trained with \method generalize to multi-app tasks in downstream evaluations.
See \Cref{app:agent-interface} for full details on the \method environment interface.

Ultimately, \method creates environments consisting of tools $\mathcal{A} = \left\{ a_1, \dots, a_A \right\} $ and tasks $\mathcal{T} = \left\{ T_1, \dots , T_N \right\} $ within these apps.
The key contribution of \method is controllability of the composition and properties of the task set $(\mathcal{A}, \mathcal{T})$ via the following controls:
\begin{itemize}[itemsep=0pt,topsep=0pt,parsep=0pt,partopsep=0pt,parsep=0pt,leftmargin=*]
  \item $A$ \textbf{number of apps}: More apps means more diversity over tools. 
    When generating apps, we ensure an equal balance across the 8 taxonomy categories previously defined.
  \item $T$ \textbf{number of tasks per-app}: This is the number of tasks generated based on functionalities within a particular app. A higher $T$ means more diversity of tasks within an app.
  \item $S$ \textbf{number of initial databases per-task instruction}. We can keep the same task instruction, verifier, and oracle solver but vary the initial database state $\text{Db}^{0}$ to form different tasks.
  \item $D$ \textbf{amount of distractor data} per initial database state. This is the total number of data batches, including the core batch $\text{Db}_{\text{core}}^{0}$, that make up the initial database state.
\end{itemize}
The task set diversity is characterized by $A,T,S$, producing $N = A\cdot T\cdot S$ total tasks.
The data complexity is characterized by $D$, which does not change the number of tasks.

\begin{figure}[t]
    \centering
    \begin{subfigure}[b]{0.32\linewidth}
        \centering
        \includegraphics[width=\linewidth]{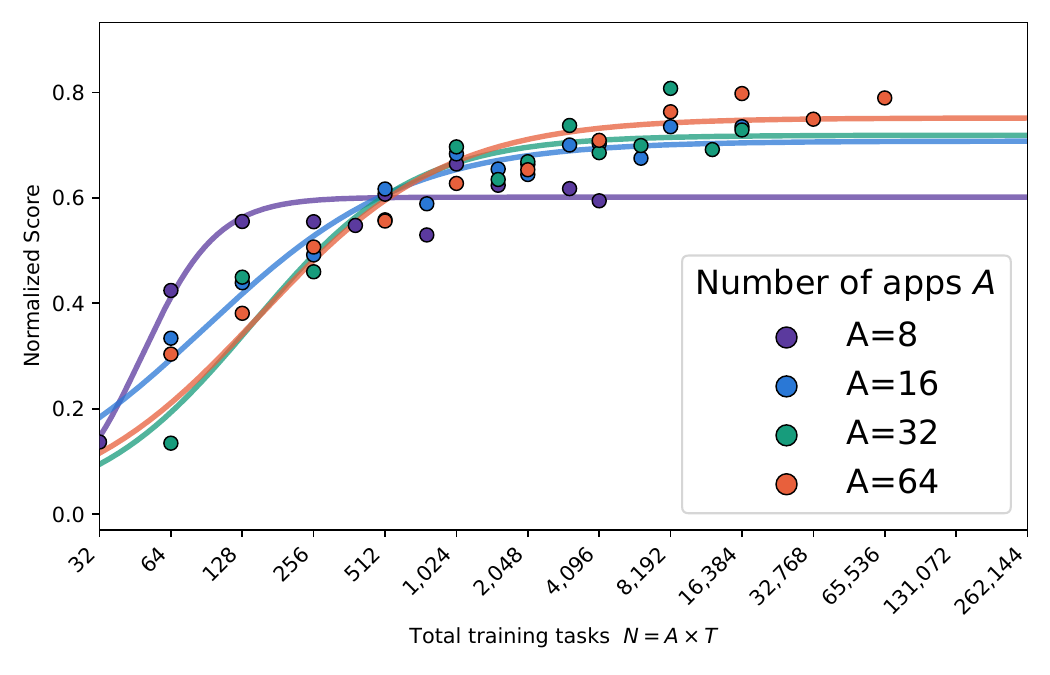}
        \caption{App vs.\ task allocation}
        \label{fig:apps-vs-tasks-main}
    \end{subfigure}
    \hfill
    \begin{subfigure}[b]{0.32\linewidth}
        \centering
        \includegraphics[width=\linewidth]{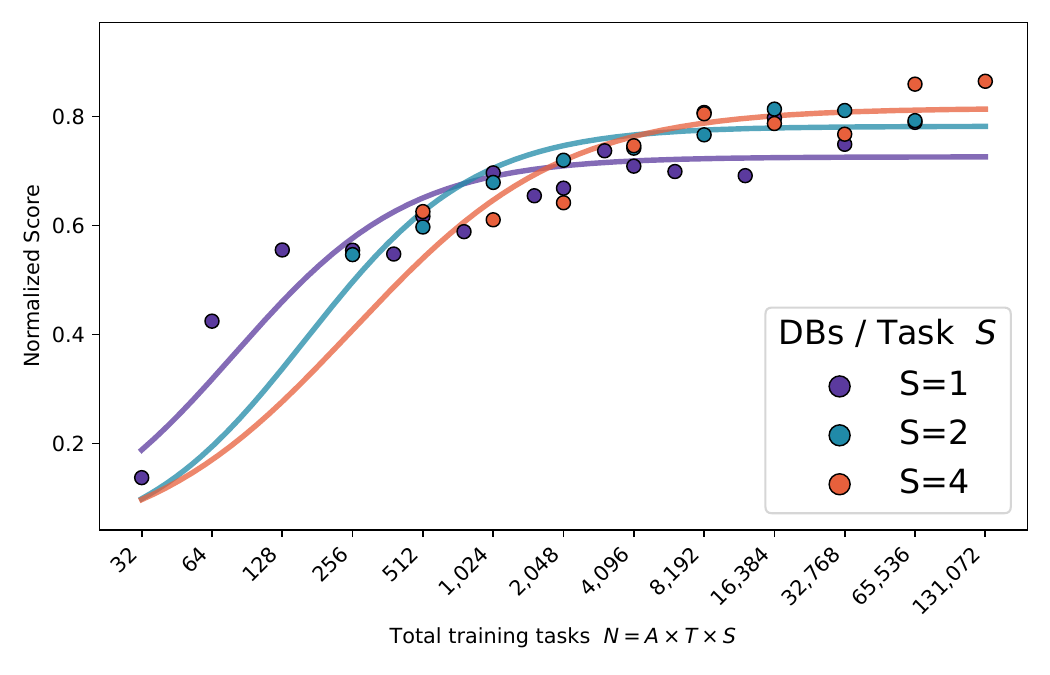}
        \caption{Database breadth}
        \label{fig:database-init-main}
    \end{subfigure}
    \hfill
    \begin{subfigure}[b]{0.32\linewidth}
        \centering
        \includegraphics[width=\linewidth]{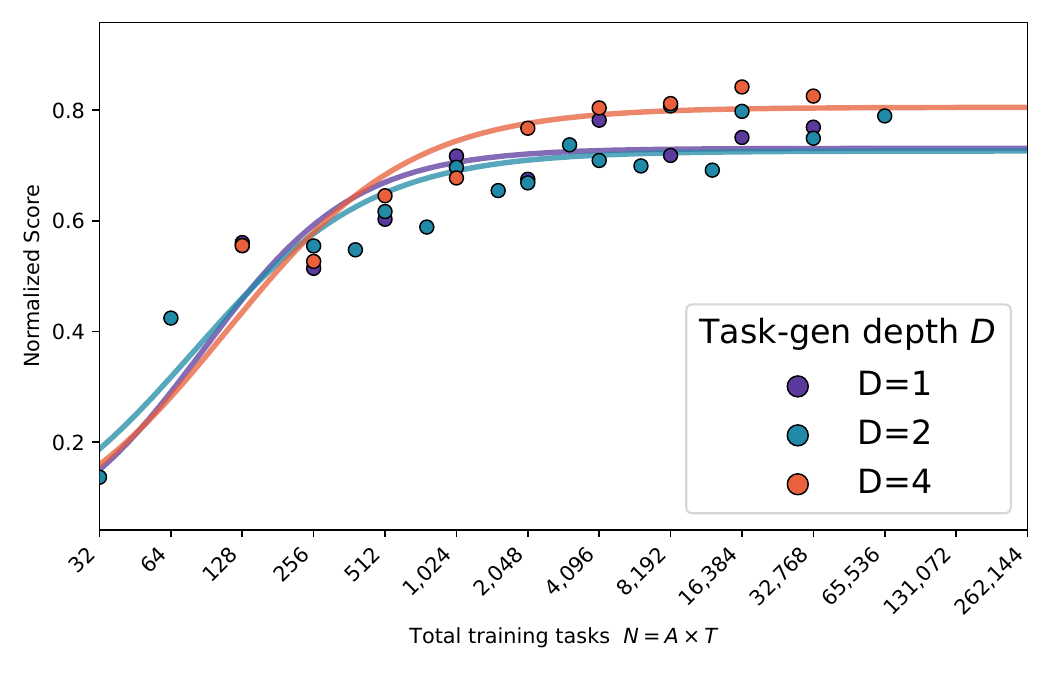}
        \caption{Database depth}
        \label{fig:database-depth-main}
    \end{subfigure}
    \caption{
    Effect of app-to-task allocation, database breadth, and database depth on the 9B student.
    The x-axis shows the number of tasks $N$, and the y-axis shows aggregate downstream performance.
    \textbf{(a)} Curves correspond to different allocations of $A$ and $T$ at a fixed total task count $N=AT$; at a fixed value of $N$, the highest curve identifies the best allocation.
    \textbf{(b)} Curves correspond to different numbers of database instantiations $S$ per task.
    \textbf{(c)} Curves compare minimal task state ($D=1$), moderate distractor data ($D=2$), and a larger distractor set ($D=4$).
    }
    \label{fig-allocation-main}
\end{figure}

\section{Experiments}
\label{sec:experiments}

\subsection{Experimental Setup}
\label{sec:experimental-setup}

We now conduct an analysis to determine the relationship between the \method controls described in \Cref{sec-control} and how agents trained on the generated data generalize to downstream agent benchmarks.
We thus use \method to generate many task sets $\left\{  (\mathcal{A}_i, \mathcal{T}_i) \right\}$ based on different sets of apps and then train an agent on each task set independently.
In total we generate 458,752 tasks, across 64 apps which consist of 10,472 APIs with 6,266 write APIs and 4,206 read APIs.
The median number of APIs per app is 144 and it ranges per app from a min of 51 to a max of 443.
The generated APIs take between 0 and 12 input arguments. 
The median number of data class definitions per database schema implementation is 55 ranging from 10 to 135.
Apps have between 31 and 143 functionalities with a median of 81.
The median number of functionalities per generated task is 2 ranging from 1 to 7.
The core database state which does not contain the extra distractor data $\text{Db}_{\text{core}}^{0}$ has a median of 59 rows.
See \Cref{sec-method-stats} for more statistics and full histograms about the generated tasks and apps.

Each episode includes a binary verifier, so agents can be directly trained in \method via reinforcement learning with verifiable rewards.
However, running an RL loop for every setting in our analysis is prohibitively expensive, so we instead collect trajectories once and then use supervised fine-tuning (SFT) to train on subsets $\mathcal{T}_i$ at the given control settings.
We use GLM-5.2~\citep{glm5team2026glm5} as the expert and allow it up to four attempts per episode.
We retain the first attempt for which the verifier indicates successful task completion and discard episodes for which none of the four attempts succeeds.
The task counts used throughout this analysis thus each have exactly one trajectory.
We distill the retained expert trajectories with the standard next-token prediction objective into student models from the Qwen3.5 family \citep{team2026qwen3}.
The loss is applied only to the reasoning and action tokens.
By default, we train models for 1 epoch over the dataset.
Training continues until the validation loss increases over a set number of updates, after which we select the checkpoint with the lowest validation loss for evaluation.
This dataset constitutes 5.9B total tokens of which 721M are agent reasoning and action output for actual supervision.
We train 849 models across the different task populations controlling for the different dataset composition properties.
\Cref{supp-training-details} lists all training details and more statistics about the trajectory dataset.

\begin{figure}[t]
    \centering
    \includegraphics[width=0.6\linewidth]{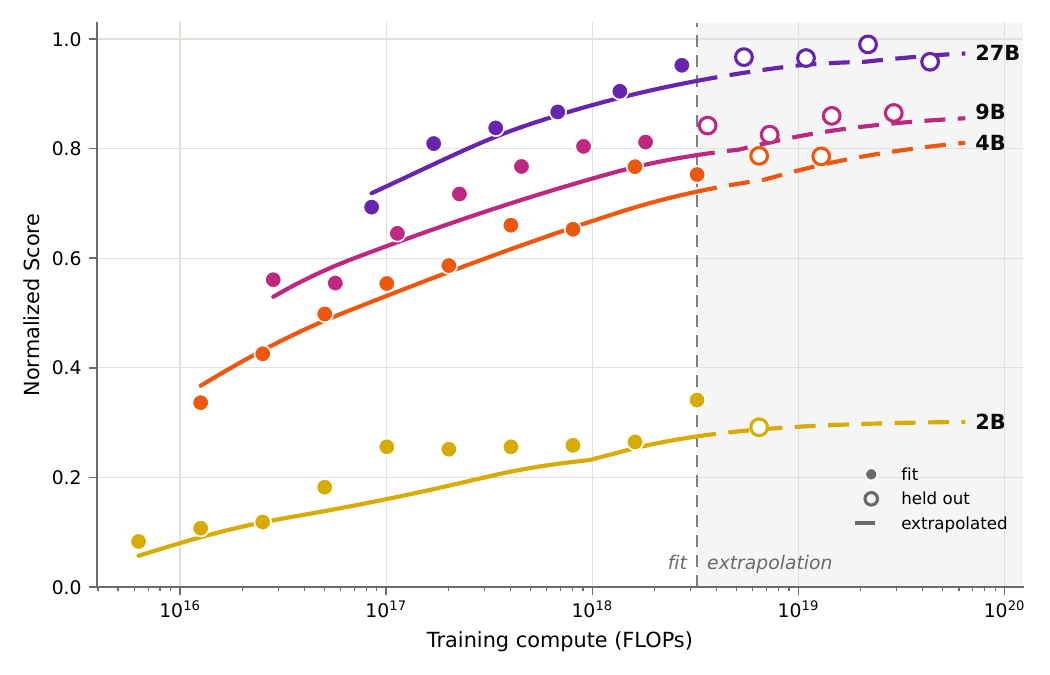}
    \vspace{-10pt}
    \caption{
        Visualization of the proposed data composition dependent scaling law and its extrapolation to higher compute budgets.
        The x-axis reports the training FLOPs of the SFT training and the y-axis is the normalized score indicating the combined generalization performance across various benchmarks.
        The points are normalized scores for models trained on an optimal setting, meaning the data setting that performs the best at that FLOP budget, representing the FLOP/performance allocation frontier.
        The lines are a visualization of the per parameter count scaling model, fit without the held-out runs in the top 25\% of training FLOPs.
    }
    \label{fig:model-size-flops}
    \vspace{-10pt}
\end{figure}

We evaluate on a representative set of benchmarks containing multi-step tool use in tools and tasks that are distinct from the \method environments we train on, thus evaluating generalization.
We focus on benchmarks where the user in the loop is not a primary challenge as \method does not generate data for these scenarios.
All tasks are evaluated based on the success rate of the agent being able to achieve the stated goal while avoiding unintended changes.
\begin{itemize}[itemsep=0pt,topsep=0pt,parsep=0pt,partopsep=0pt,parsep=0pt,leftmargin=*]
    \item \textbf{AppWorld} \citep{trivedi2024appworld} evaluates agents in device assistant tasks across 9 apps and 457 APIs.
        Tasks involve complex user data and combining information across apps in the face of distractors and constraints.
        We evaluate on the Test Normal (Test-N) set of 168 tasks and Test Challenge (Test-C) set of 417 tasks.

    \item \textbf{BFCLv3 Multi-Turn} \citep{patil2025bfcl} has 129 APIs across 8 domains. We restrict to the multi-turn split which has 800 tasks which also has splits for explicitly testing robustness to missing APIs, missing parameters, and long tool outputs.

    \item \textbf{ASTRA-Bench} \citep{xiu2026astrabench} focuses on persona-driven evaluation, where each task combines rich user context with 27 APIs across 6 apps. We evaluate on the most complex ``High'' setting consisting of 939 tasks.

    \item \textbf{\sapihold.} We construct a held-out set of new apps, tasks and data profiles using \method allowing us to measure within-distribution cross-app generalization. We generate 4 apps with 25 tasks per app where each task is generated with the maximum data complexity setting.
\end{itemize}

We use a subset of these benchmarks, which we call the \emph{diagnostic set}, for the scaling and data composition analysis.
AppWorld and \sapihold are used as the diagnostic set because evaluating on these benchmarks alone required 1.8 million episodes.
Because raw scores are not directly comparable across benchmarks, we normalize the model's score when doing analysis on the diagnostic set.
To do so, for benchmark $i$ and model size $M$, let $\theta_0^{M}$ be the starting model parameters, $\Theta_i$ the population of all trained models across all sizes and $\text{Succ}_i(\theta)$ the success rate of model parameterized by $\theta$ averaged over 4 trials per episode.
We then compute the normalized score of model $\theta^{M}$ as $P(\theta^{M})=\frac{1}{K}\sum_{i=1}^K \frac{\text{Succ}_i(\theta^{M}) - \text{Succ}_i(\theta_0^{M})}{\max_{\theta' \in \Theta_i}\text{Succ}_i(\theta') - \text{Succ}_i(\theta_0^{M})}$.
$P$ is thus bounded between $[0,1]$ starting at the zero-shot performance of the Qwen student without \method training and ending at the maximum observed training performance on this benchmark.
\Cref{supp-benchmark} contains complete benchmark details and a full description of the evaluation protocol.

\begin{table}[t]
    \centering
    \resizebox{\linewidth}{!}{%
    \begin{tabular}{lcccccccccccc}
        \toprule
        & \multicolumn{3}{c}{AppWorld} & \multicolumn{3}{c}{ASTRA-Bench} & \multicolumn{3}{c}{BFCL} & \multicolumn{3}{c}{\method-Holdout} \\
        \cmidrule(lr){2-4} \cmidrule(lr){5-7} \cmidrule(lr){8-10} \cmidrule(lr){11-13}
        Approach & 4B & 9B & 27B & 4B & 9B & 27B & 4B & 9B & 27B & 4B & 9B & 27B \\
        \midrule
        Zero-shot     & 14.7 & 26.4 & 61.5 & 57.4 & 71.8 & 80.2 & 27.3 & 39.3 & 54.4 & 0.3 & 1.5 & 20.0 \\
        Nemotron      & 17.5 & 20.8 & -    & 54.4 & 67.7 & - & 15.9 & 39.7 & - & 1.0 & 0.5 & - \\
        \sapiSmallSub & 38.4 & 53.0 & 73.8 & 71.1 & 74.2 & 77.3 & 44.9 & 47.5 & 59.6 & 51.3 & 56.0 & 64.0 \\
        \sapiSmallOpt & 45.0 & 56.2 & 74.1 & 72.3 & 75.5 & 79.3 & 48.8 & 49.9 & 61.5 & \textbf{56.0} & 63.0 & 68.0 \\
        \sapiFull     & \textbf{66.4} & \textbf{77.0} & \textbf{81.5} & \textbf{74.2} & \textbf{76.8} & \textbf{82.3} & \textbf{51.1} & \textbf{53.6} & \textbf{65.0} & 53.0 & \textbf{64.0} & \textbf{69.8} \\
        \bottomrule
    \end{tabular}%
    }
    \caption{
        Unnormalized success rates on downstream tool-use agent benchmarks. AppWorld reports the Test-C split (see \Cref{tab:appworld-breakdown} for both splits). \method rows are trained with \method data; ``-Opt'' uses the optimal data composition from our scaling analysis, and ``-Big'' is a scaled-up version. 
    }
    \label{tab:baseline-comparison}
\end{table}

\subsection{Task Properties Analysis}
\label{sec-task-properties}

We first vary the individual \method task-generation controls described in \Cref{sec-control} under fixed task or data budgets and evaluate downstream generalization on the diagnostic set.

\textbf{Tasks vs. apps}:
First, we study how to allocate a fixed task budget between generating more apps, and thus more APIs, or generating more tasks for existing apps.
The total task count is $N = AT$, where $A$ is the number of apps and $T$ is the number of tasks per app.
For this comparison, we fix the number of databases per task at $S=1$ and the distractor-data setting at $D=2$.
We sweep all combinations of $A \in \left\{ 8,16,\dots,64 \right\}$ and $T \in \left\{ 8,12,16,\dots,1024 \right\}$, training a separate model on the tasks generated for each configuration.
We evaluate each trained model $\pi_\theta$ on the diagnostic benchmarks to obtain $P(\theta)$.
For each app count $A$, we model the relationship between $P$ and $N$ as $T$ varies.
We fit the following generalized logistic model to the observed data using least-squares minimization, estimating parameters $U,b,b_N,\nu$ separately for each model size $M$ and app count $A$ where $\sigma$ is the sigmoid function.
\begin{equation}
    \hat{P}_{M,A}(N) = U \cdot \sigma\left( b + b_N \log_2 N \right)^{\nu}
    \label{eq-fit-local}
\end{equation}

\Cref{fig:apps-vs-tasks-main} compares allocations of $A$ and $T$ and their fitted curves across training task budgets $N$ for the 9B model.
At lower task budgets, fewer apps yield better performance.
We hypothesize that concentrating tasks within a smaller set of tools and data structures yields larger initial gains.
As the task budget increases, however, a larger number of apps becomes preferable.
Specifically, $A=8$ performs best at $64$--$512$ tasks and $A=64$ is clearly the best starting at $4096$ tasks.
The fitted performance ceiling also increases with $A$, with $A=8$ saturating at a substantially lower value than $A=64$.
The gains from increasing app diversity are not due to tasks exhausting each app's functionality.
Even at $T=1024$ tasks per app, the median fraction of tools used is only $51\%$, reflecting the complexity of \method apps in their tool counts and input-argument counts and types (\Cref{sec-method-stats}).
Prior work such as \citet{tu2026scaleenv} also finds benefits from scaling the number of tools but does not study how the allocation changes with task budget.
\Cref{supp-analysis} shows the same trends across the 2B, 4B, and 27B model sizes.

\textbf{Data profiles vs. tasks:} We then study database breadth by varying $S$, the number of database instantiations generated per task instruction.
Each database instantiation keeps the instruction, but changes the starting database state.
The x-axis shows the total number of training instances $N$, and each curve corresponds to a different value of $S$.
For each budget $N$, we use the optimal app-to-task allocation from the previous study, adjusted for the selected $S$ so that $N = T \cdot A \cdot S$.
Each task instruction receives its own distinct data profiles.
We fit the model in \cref{eq-fit-local} separately for each database count $S$ in place of app count $A$, yielding $\hat{P}_{M,S}$.
The results in \Cref{fig:database-init-main} show that higher task budgets favor greater database breadth.
At budgets of $32$--$512$ training instances, one data profile per instruction performs best; at $512$--$4096$, two profiles perform best; and beyond $4096$, four profiles perform best.
The fitted performance ceiling also increases with $S$.
We attribute these gains to learning to solve the same task across different initial database states, since data manipulation is a key challenge in our training tasks.
This suggests that, as the budget grows, generating additional data profiles for existing instructions becomes preferable to adding new instructions with new data profiles.

\textbf{Effect of data complexity:}
We also study database depth, which controls the amount of distractor data within each data profile.
We compare $D=1$, which includes only the minimum state needed to solve the task; $D=2$, which adds a moderate amount of distractor data; and $D=4$, which adds a larger distractor set.
As before, the x-axis shows the task budget $N$, the y-axis shows aggregate downstream performance, and each curve corresponds to a different $D$; we use the optimal allocation of $A$ and $T$ with $S=1$.
\Cref{fig:database-depth-main} shows a large improvement from $D=1,2$ to $D=4$ at high task budgets, indicating that training with distractor data improves downstream scaling.
This pattern may partly reflect the current downstream benchmarks' limited sensitivity to increasingly difficult database states.
Benchmarks that place greater emphasis on retrieval and decision-making over dense database states may exhibit different scaling behavior along this dimension.

\subsection{Scaling Law}
\label{sec-scaling}

In this section, we take the learned lessons from the previous section to now fit a predictable relationship between all of the \method controls and the diagnostic set performance.
In total, we use all 849 trained models across all the different dataset populations which pairs each setting of $(A,T,S,D)$ with downstream normalized score across each model size $M$.
Guided by the previous analysis, we formulate the scaling law equation to build off of \Cref{eq-fit-local} but now fit a data composition dependent upper limit $U(A,S,D)$ rather than the previous fixed $U$ value.
Specifically, we use the following form, where $c_0, c_A, b_0, b_T, \nu$ are fit from the data using least-squares minimization.
\begin{equation}
    \widehat{P}_M^{1}(A,T,S,D) = \sigma(c_0 + c_A \log_2 (A \cdot S \cdot D)) \sigma(b_0 + b_T \log_2 T)^{\nu}
    \label{eq:appgen-scaling-law}
\end{equation}
We also compare to a form that only uses the total number of trajectories $\hat{P}_M^{0}(N) = U \sigma(b_0 + b_N \log_2 N)^{\nu}$.
We measure the FLOPs of each training run with the standard Chinchilla approximation $C \approx 6ML$ \citep{hoffmann2022training} where $L$ is the number of total supervision tokens in the training trajectories.
We hold out the runs in the top 25\% of training FLOPs for extrapolation and fit only on the remaining runs, thus evaluating a 16x extrapolation increase in FLOP budget.

\Cref{fig:model-size-flops} visualizes the data composition dependent scaling curve across each model size along with the actual observed frontier optimal data points at various FLOP budgets.
We only display the optimal points at the given FLOP budgets, but the curves are fit to all the data points.
The data composition aware scaling law achieves a held-out mean absolute error (MAE) value of $0.0285$ in the extrapolation region (the top 25\% of training FLOPs) compared with an MAE of $0.1706$ for the data-quantity-only scaling law $\hat{P}_M^{0}$.
\Cref{supp-alloc-analysis} reports the fit quality for each model size.
The findings are also the same if $\hat{P}_M^{0}$ is parameterized by the token count, and equivalently the FLOPs, instead of the number of tasks $N$.
This demonstrates that the proper composition of the dataset is important for agent scaling rather than just sheer scale.
The data dependent scaling law shows that the upper ceiling of performance is increased by increasing the app count, number of data profiles per task and data profile complexity with the number of tasks per app controlling the rate of progress towards this maximum.
Furthermore, we demonstrate it is possible to extrapolate to predict agent performance under up to $16\times$ larger compute budgets.

\begin{table}[t]
    \centering
    \footnotesize
    \begin{minipage}[t]{0.40\linewidth}
        \centering
        \begin{tabular}{lcc}
            \toprule
            \textbf{BFCL} & 4B & 8B \\
            \midrule
            Zero-shot         & 33.5 & 41.3 \\
            Agent World Model & 40.8 & 42.3 \\
            EnvScaler         & 45.0 & 51.9 \\
            EnvFactory        & 48.5 & 49.0 \\
            \sapiFull         & \textbf{50.5} & \textbf{55.6} \\
            \bottomrule
        \end{tabular}
    \end{minipage}
    \hfill
    \begin{minipage}[t]{0.56\linewidth}
        \centering
        \begin{tabular}{lcccc}
            \toprule
            & \multicolumn{2}{c}{4B} & \multicolumn{2}{c}{9B} \\
            \cmidrule(lr){2-3} \cmidrule(lr){4-5}
            \textbf{AppWorld} & Test-N & Test-C & Test-N & Test-C \\
            \midrule
            Zero-shot      & 47.5 & 14.7 & 45.4 & 26.4 \\
            ToolACE        & 20.2 & 10.9 & 18.6 & 12.4 \\
            ToolAlpaca     & 25.7 & 17.3 & 24.4 & 15.7 \\
            SimSAT         & 66.8 & 58.8 & 74.5 & 65.4 \\
            \sapiFull      & \textbf{81.0} & \textbf{66.4} & \textbf{83.3} & \textbf{77.0} \\
            \bottomrule
        \end{tabular}
    \end{minipage}
    \vspace{-5pt}
    \caption{
        Comparison to per-benchmark baselines from prior work on BFCL (left) and AppWorld (right) showing success rates. 
        BFCL results are on Qwen3 and AppWorld results are on Qwen3.5.
    }
    \label{tab:prior-work-baselines}
    \vspace{-10pt}
\end{table}

\subsection{Baseline Comparison}
\label{sec-baseline} 

Next, we demonstrate the performance of \method across all the downstream benchmarks and compare to prior works in task synthesis.
For this comparison, we also include performance on the ASTRA-Bench and BFCLv3 benchmarks.
First, to show the value of the FLOP optimal training data composition settings derived from the previous scaling law analysis we compare to two versions of models trained with \method data and a large scale existing tool-use agent dataset:
\begin{itemize}[itemsep=0pt,topsep=0pt,parsep=0pt,partopsep=0pt,parsep=0pt,leftmargin=*]
    \item \sapiSmallSub: We train one model on 32,768 \method tasks, but we ignore the lessons from the previous sections and instead choose $A=32$, $T=1024$, $S=1$, $D=2$.
    \item \sapiSmallOpt: We train under the same task budget as the above, but now maximize app and database diversity by reducing task prompts, with $A=64$, $T=128$, $S=4$, $D=4$.
    \item Nemotron \citep{nvidia2025nemotronagentic}: We train on the full Nemotron-Agentic dataset.
    \item \sapiFull: We train on the full \method generated dataset but now for 4 epochs.
\end{itemize}

The results from \Cref{tab:baseline-comparison} show the performance of these approaches across all benchmarks.
We omit the 2B size due to its far worse performance in the previous section.
First comparing \sapiSmallOpt and \sapiSmallSub, we see the value of the right data recipe with increasing $S$, $D$ and $A$ while keeping the total task budget fixed.
\sapiSmallOpt does consistently better than \sapiSmallSub, with especially large deltas in \sapihold and the 4B model.
When scaling up to \sapiFull, results further improve on most benchmarks with especially large gains in AppWorld.
This method greatly outperforms Nemotron, which contains relatively simple and short-horizon data that does not generalize to the complexity of these benchmarks and even underperforms the zero-shot starting model in 5 of the 8 settings.

Next, we compare with results reported in prior work in \Cref{tab:prior-work-baselines}.
On AppWorld, we compare with \textbf{ToolACE} \citep{liu2024toolace} and \textbf{ToolAlpaca} \citep{tang2023toolalpaca}, which synthesize tool-use demonstrations with LLM-simulated API responses, and \textbf{SimSAT} \citep{lee2026environment}, which uses LLM simulation to synthesize long-horizon trajectories.
On BFCL, we compare with \textbf{Agent World Model} \citep{wang2026agentworldmodel}, which synthesizes database-backed tool environments; \textbf{EnvScaler} \citep{song2026envscaler}, which generates environment skeletons and verifiable task scenarios; and \textbf{EnvFactory} \citep{xu2026envfactory}, which builds executable environments from real-world resources.
Since these BFCL baselines only report results for Qwen3, we also train \sapiFull on Qwen3-4B/8B.

ToolACE and ToolAlpaca generally underperform the zero-shot models on AppWorld, indicating limited transfer to its complex stateful tasks.
SimSAT, which uses LLM simulation of long-horizon interactions and GLM-based distillation, provides a substantially stronger baseline.
\sapiFull achieves the highest scores on both AppWorld splits, reaching $66.4\%$ and $77.0\%$ on Test-C at 4B and 9B, respectively, exceeding SimSAT by $7.6$ and $11.6$ percentage points.
On BFCL, the prior executable environment synthesis methods all improve over zero-shot, while \sapiFull further surpasses the strongest baseline by $2.0$ percentage points at 4B (EnvFactory) and $3.7$ points at 8B (EnvScaler).
However, their training regimes differ substantially with these methods combining SFT and RL on smaller datasets, whereas \sapiFull uses SFT on a much larger synthetic corpus.
These comparisons therefore do not isolate the effect of the synthesis method or establish relative data efficiency.
They do demonstrate that \method supports large-scale synthesis of tasks that enable strong downstream generalization. 
The strong results overall show how our scaling laws can guide the construction of effective training datasets. 
See \Cref{app:additional-results} for more results on transferring findings to other datasets, evaluation breakdowns, and further scaling law analysis.

\section{Conclusion}
\label{sec:conclusion}

This work introduces \method, a controllable pipeline for synthesizing executable environments for training and studying tool-use agents.
\method uses LLMs to generate executable environments with a structured database schema and Python API implementations.
\method then generates verifiable and feasible tasks with a starting database state.
This design allows us to independently scale the diversity of tools, data profiles, and tasks.
We empirically show that the composition of training data substantially affects downstream performance.
We combine these results in a joint scaling law that predicts downstream performance and gives optimal data composition settings for a given compute budget. 
A limitation of \method is that it does not simulate a user in the loop that interacts with the agent.
Such a setting is important for realistic settings where agents need to query a user to ask for additional details or clarifying questions.
Furthermore, \method focuses less on information retrieval tasks involving long and complex documents in realistic user settings like long email chains or note documents.
Since all content in \method is synthesized from scratch by the LLM it lacks this semantic depth.
Furthermore, the \method scaling law is only validated on a single data pipeline and it is not guaranteed to transfer to different data pipelines or training settings.

All tool-use experiments in this work are conducted in research-only simulated environments. The APIs, tasks, datasets, verifier infrastructure, and training procedures described here do not represent or imply any deployed Apple product, production system, or product roadmap.

\bibliography{ref}
\bibliographystyle{iclr2027_conference}

\newpage
\appendix
\crefalias{section}{appendix}
\crefalias{Section}{Appendix}

\startcontents[appendix]
\printcontents[appendix]{}{1}{\section*{Appendix Contents}\vspace{5mm}}

\section{\method---Full Details}

\subsection{App Generation Details}
\label{supp-app-gen} 

\begin{promptbox}[label=app:appgen-functionality-generation-prompt]{Functionality generation prompt. Sent to the builder LLM to generate the functionalities of a given app, which are later sampled to construct task instructions.}
Give me every possible thing I can do in the {app} app, be comprehensive. Do not include any functionality that requires interacting with another app (for example, importing data from another app). Functionality that can be done entirely in-app is fine (for example, managing connections within {app} itself). Focus on the core, domain-specific things a user does in {app} (for example, the products, content, or services the app is actually about). Do not produce functionality that is solely about account management (for example, signing up, logging in, changing your password, updating email or notification preferences, or managing account settings).
\end{promptbox}

\begin{promptbox}[label=app:appgen-database-schema-prompt]{Database schema generation prompt. Used to generate an app's database schema (\texttt{db.py}) as a collection of Python data classes, given the app metadata: the app name, its description, and the functionality list produced by the preceding prompt.}
I want to create a data model of objects in a relational database. Specifically, I want to mock up a version of the {app} app with data models that can back any possible functionality in the app. I want this expressed as a series of Python dataclass objects. I want an exhaustive description of all possible data I would need for every possible interaction in the {app} app. Write all of the dataclasses into db.py and do not stop until every possible data type is covered. Make sure the data classes cover all of the functionalities listed below.

Functionalities:
```
{functionalities}
```
\end{promptbox}
The generation here uses the builder LLM, which we use Gemini-3.1 Pro for.
In the app creation, we use Gemini-3.1 Pro through the Gemini CLI harness.
This allows the model to agentically create the various app components and take on more complex changes.
The Gemini CLI harness is only used for the app creation and not for the task generation.

\begin{promptbox}[label=app:appgen-api-generation-prompt]{API generation prompt. Used to generate an app's API implementation as ordinary Python functions, conditioned on the previously generated database schema.}
I want to mock up all of the APIs in {app}. Implement all of the APIs in apis.py. To build out the APIs you are only allowed to use the four CRUD functions the database exposes, and you must build APIs using all of the data models in db.py. Your APIs must cover any possible task across the functionalities listed below. Decorate every function that does not require accessing the current user (for example searches and public catalog lookups) with the `no_auth` decorator. Functions that need the current user must not use this decorator.

Data models:
```
{db_models}
```

Functionalities:
```
{functionalities}
```
\end{promptbox}
When implementing the APIs, we only allow an API to interact with the database state via one of four defined create, read, update, delete functions.
We use this access gating so we can directly track all changes made to the database and keep this access log in the database object.

\subsection{Task Generation Details}
\label{supp-task-gen} 

\begin{promptbox}[label=app:appgen-task-generation-prompt]{Task generation prompt. Used to generate unambiguous task instructions from a sampled subset of an app's functionalities.}
You are generating task instructions for an AI agent that interacts with the {app} app via APIs.

Below are functionalities of the app:
{functionalities}

Existing tasks (for reference, do NOT duplicate these):
{existing_tasks}

Generate exactly {n_tasks} diverse, concrete task instructions that exercise the functionalities above. Each task should:
1. Be SHORT: a single, clear instruction phrased the way a real person casually speaks it, in 1-2 sentences (roughly 15-40 words), NOT a long multi-step plan.
2. Explicitly mention the app name "{app}" in the task instruction
3. Include specific product names, quantities, or criteria where appropriate
4. Vary in difficulty (some simple, some multi-step)
5. Not duplicate any existing task
6. Be an actionable instruction (e.g. "add X to my cart", "place an order for Y"), NOT a question that requires an answer (e.g. "what is the price of X?", "how many Y are in stock?"). The agent should be performing actions, not retrieving information to answer the user.{additional_rules}

Keep tasks SHORT - the difficulty should come from a single constraint or decision the agent must reason about, not from stacking many steps.

Output your tasks as a Python list of strings:
```python
tasks = [
    "...",
    "...",
]
```
\end{promptbox}

\begin{promptbox}[label=app:appgen-initial-state-prompt]{Initial state generation prompt. Used to instantiate the core initial database state containing the records and relations necessary to make a task instruction meaningful and solvable.}
Using the defined data models, write a setup_database function that returns a list of dataclass instances to populate the database with initial data. The data should:
1. Include all entities needed so that the task instruction below can be successfully completed
2. Include some distractor data (extra realistic entries that aren't directly needed for the task) to make the scenario more realistic
3. Use consistent IDs and foreign key references between related entities
4. Use realistic values (names, emails, prices, etc.)
5. Any user-related data (User, Address, etc.) for the main user MUST use the exact values shown in the "Main user" section below.

The output should be in format:
```python
def setup_database() -> list:
    ...
```
Do not output any other code. Do not include any import statements and reference the data models exactly as they are named.

Data models:
```
{db_models}
```

APIs (for reference on available operations):
```
{api_docs}
```
\end{promptbox}
The actual database state is thus just a list of instantiated dataclass objects.
We also couple the database with a change log of executed create, update, read or delete operations that happened on the database.

\begin{promptbox}[label=app:appgen-verifier-generation-prompt]{Verifier generation prompt. Used to generate the programmatic verifier that determines whether a database state satisfies the task instruction.}
Using the defined APIs, write a verifier for the below "Instruction" to check if the task is successfully completed. The output should be in format:
```python
def verify(db: Database, answer: str | None = None) -> bool:
    ...
```
Do not output any other code. Do not include any import statements and reference the APIs exactly as they are named.

APIs:
```
{api_docs}
```
Instruction: `{instruction}`
\end{promptbox}

\begin{promptbox}[label=app:appgen-oracle-generation-prompt]{Oracle generation prompt. Used to generate the oracle solver program that completes the task instruction by calling app APIs from the initial database state.}
Using the defined APIs, write a solve function that performs the actions described in the instruction to complete the task. The solve function receives a Database object that has been pre-populated with initial data.

Hard requirements:
- Only call the public APIs listed below to make changes. Do not call, or read attributes from, anything whose name starts with `_` (private/internal helpers are off-limits).
- Do not call `create_data_model`, `read_data_model`, `update_data_model`, or `delete_data_model` directly, do not use `setattr`, and do not read or write `db` (e.g. `db.data`, `db.<field> = ...`, `db.<field>.append(...)`) directly. Every change to the database must go through a public API call.

The output should be in format:
```python
def solve(db: Database) -> None:
    ...
```
Do not output any other code. Do not include any import statements and reference the APIs exactly as they are named.

Data models:
```
{db_models}
```

APIs:
```
{api_docs}
```

solve_task definition (for context on how solve is called):
```
{solve_task_def}
```
Instruction: `{instruction}`

Task file (for context on the setup data and verifier):
```
{task_file}
```
\end{promptbox}

\begin{promptbox}[label=app:appgen-database-expansion-prompt]{Database expansion prompt. Used to expand the core initial database state with distractor records that increase retrieval and selection difficulty without changing the intended solution.}
Using the defined data models, write a setup_additional_stage_{stage_n} function that returns a list of dataclass instances to add EXTRA distractor data on top of the data already produced by setup_database AND by the earlier distractor stages (setup_additional_stage_1 .. setup_additional_stage_{prev_n}). The goal is to make the task harder by introducing realistic but irrelevant entries (e.g. additional users, similar-looking orders, products in the same category, etc.).

This is stage {stage_n} in a sequence of distractor stages. When the task runs, the records returned by setup_database and by EVERY setup_additional stage (1 through {stage_n}) are all loaded into the SAME database together. Your stage {stage_n} data must therefore coexist with everything the earlier stages already added. To keep the overall distractor set diverse, your stage {stage_n} data SHOULD be meaningfully DIFFERENT from what the earlier stages produced (shown in the "Earlier setup_additional stages" section below) - introduce new kinds of realistic-but-irrelevant records rather than repeating the same patterns, entities, or IDs.

Hard requirements:
1. The additional data MUST NOT cause the verifier to return True before the solve runs (the task must still start unsolved).
2. The additional data MUST NOT prevent the solve function from completing successfully (the task must still be solvable).
3. Use IDs that do not collide with anything already created in setup_database OR in ANY earlier setup_additional stage.
4. The additional data SHOULD be relevant to the main user (shown in the "Main user" section below) - e.g. extra orders, addresses, payment methods, reviews, browsing history, or other records belonging to or associated with the main user - so the distractors are plausible noise in the main user's account. These records must NOT interfere with the solvability of the main task or cause the verifier to pass prematurely.

The output should be in format:
```python
def setup_additional_stage_{stage_n}() -> list:
    ...
```
Do not output any other code. Do not include any import statements and reference the data models exactly as they are named.

Data models:
```
{db_models}
```

APIs (for reference on available operations):
```
{api_docs}
```
\end{promptbox}

\subsection{Agent Interface Details}
\label{app:agent-interface}

Each task is a Gymnasium environment \citep{towers2024gymnasium} whose state is the instantiated database and whose transitions are app API calls.
The environment is deterministic, so a trajectory is fully determined by $\text{Db}^{0}$ and the agent's actions.

We follow an action format similar to the AppWorld environment \citep{trivedi2024appworld}: an action is a single fenced Python code block.
It executes in a Python session that persists for the whole episode, so variables defined on one turn are available on later turns.
The app API functions and the episode's \texttt{Database} object are bound in the session namespace, so the agent calls tools as ordinary Python functions.
A single turn can therefore call several tools, loop over the records returned by an earlier call, and branch on intermediate values.
Text outside the code block is treated as reasoning and is not executed.
If a turn contains no parseable code block, the environment returns a format error and the turn counts against the step limit.

The observation is the output of the previous code block execution, with tool results returned in JSON format.
When return values are \texttt{db.py} dataclasses, or lists of them, we serialize them field by field into JSON so that field names and values are explicitly displayed to the agent rather than generic class identifiers.
In other words, when an API returns an instantiated dataclass, the agent sees its full data contents.
Errors and exceptions do not end the episode.
Instead, the error message is displayed and the agent can fix the problem on the next turn.
Observations are truncated to 2,048 tokens with an explicit \texttt{[truncated]} marker.

We also expose several ``meta-tools'' which are always present in the agent's toolbox.
The meta-tools are used to discover and look up documentation for the dynamic per episode tool set the agent has to deal with.
\begin{itemize}[itemsep=0pt,topsep=0pt,parsep=0pt,partopsep=0pt,leftmargin=*]
  \item \texttt{list\_apps()}: name and one-line description of each app in the episode.
  \item \texttt{list\_tools(app)}: names and one-line summaries of one app's public API functions.
  \item \texttt{get\_tool\_documentation(app, tool)}: one tool's description, argument names, types and defaults, and return type.
  \item \texttt{complete\_task(answer=None)}: ends the episode, with an optional answer for tasks that require reporting information rather than only changing state.
\end{itemize}

Episodes in \method are capped at 32 steps.
The verifier is called only when the agent invokes \texttt{complete\_task}, so a task counts as a success only if the agent calls \texttt{complete\_task} and $v$ then returns true on the resulting database state.
An agent that reaches the step limit without calling \texttt{complete\_task} fails the task, even if its actions happened to satisfy the verifier.

\Cref{app:appgen-agent-prompt} shows the initial task prompt given to the agent. Information in the template is dynamically substituted based on the starting database state and task instruction.
\begin{promptbox}[label=app:appgen-agent-prompt]{Agent prompt. Initial task prompt given to the agent as the first user message.}
I am your supervisor and you are a super intelligent AI Assistant whose job is to achieve my day-to-day tasks completely autonomously.

To do this, you will need to interact with apps using their associated APIs on my behalf. For this you will undertake a *multi-step conversation* using a python REPL environment. That is, you will write the python code and the environment will execute it and show you the result, based on which, you will write python code for the next step and so on, until you've achieved the goal. This environment will let you interact with app/s using their associated APIs on my behalf.

Here are three key tools that you need to know to get more information

# To get a list of apps that are available to you.

```python
print(list_apps())
```

# To get the list of tools under any app listed above

```python
print(list_tools(app='xyz'))
```

# To get the specification of a particular tool

```python
print(get_tool_documentation(app='xyz', tool='abc'))
```

**Key instructions**:
- Remember you can use the variables in your code in subsequent code blocks.
- ALWAYS look at tool specifications (using get_tool_documentation) before calling a tool. Never call a tool without first using get_tool_documentation.
- Write small chunks of code and only one chunk of code in every step. Make sure everything is working correctly before making any irreversible change.
- Use `for` loops when processing lists of data rather than manually handling items one by one.
- Check the environment state before proceeding. The task may start in the middle of a user session, meaning the user may have already modified the environment. Do not assume the environment is in a fresh or default state.
- Use `print` to inspect code outputs.
- Once you have completed the task, make sure to call complete_task(). If the task asked for some information, return it as the answer argument, i.e. call complete_task(answer=<answer>). Many tasks do not require an answer, so in those cases, just call complete_task() i.e. do not pass any argument.
- Output just a single Python code block.

Using these APIs, now generate code to solve the actual task:

My name is: {{ main_user.first_name }} {{ main_user.last_name }}. My personal email is {{ main_user.email }} and phone number is {{ main_user.phone_number }}. The current date is {{ main_user.session_start_time }}.
Task: {{ input_str }}
\end{promptbox}

\subsection{\method App and Task Statistics}
\label{sec-method-stats} 
\begin{figure}[t!]
  \centering
  \begin{subfigure}[b]{0.32\textwidth}
    \includegraphics[width=\textwidth]{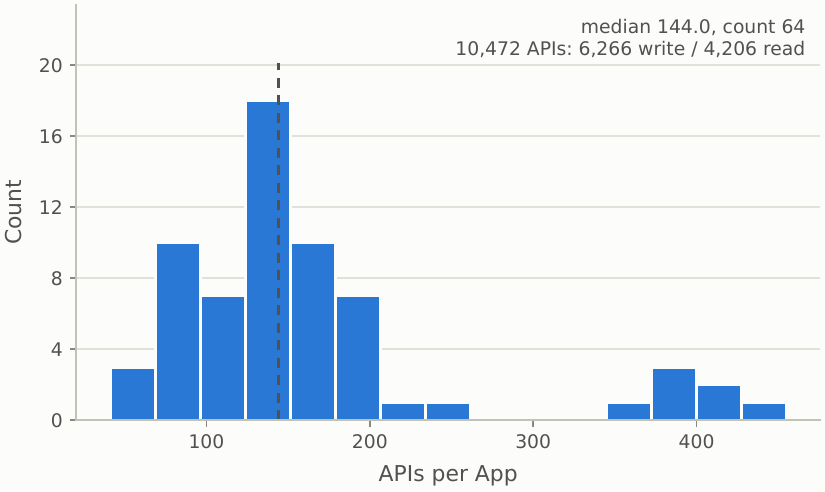}
    \caption{APIs / App}
    \label{fig-app-info-apis}
  \end{subfigure}
  \hfill
  \begin{subfigure}[b]{0.32\textwidth}
    \includegraphics[width=\textwidth]{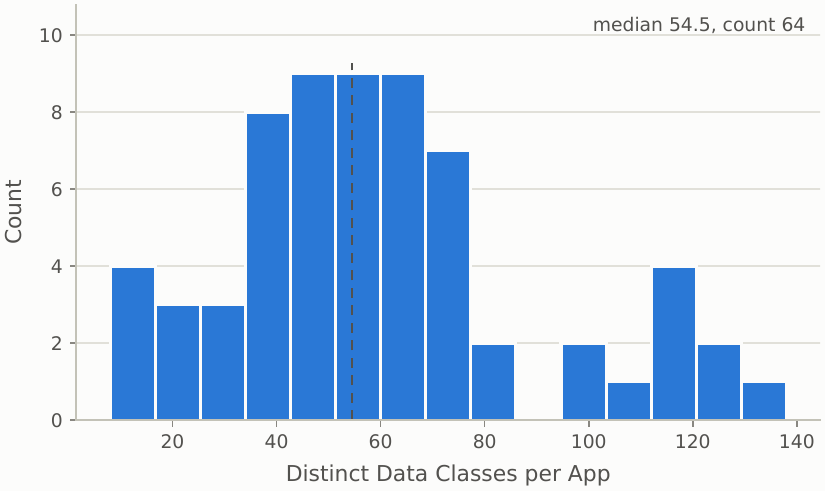}
    \caption{Data Classes / App}
    \label{fig-app-info-classes}
  \end{subfigure}
  \hfill
  \begin{subfigure}[b]{0.32\textwidth}
    \includegraphics[width=\textwidth]{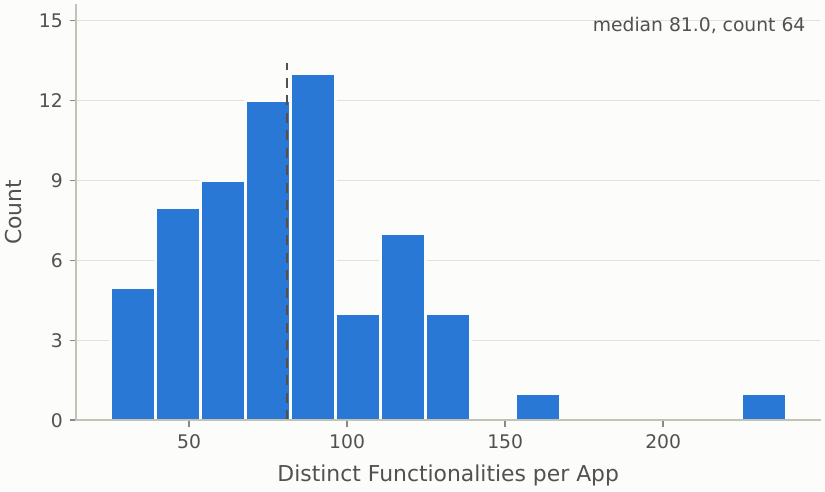}
    \caption{Functionalities / App}
    \label{fig-app-info-funcs}
  \end{subfigure}
  \\[1em]
  \begin{subfigure}[b]{0.32\textwidth}
    \includegraphics[width=\textwidth]{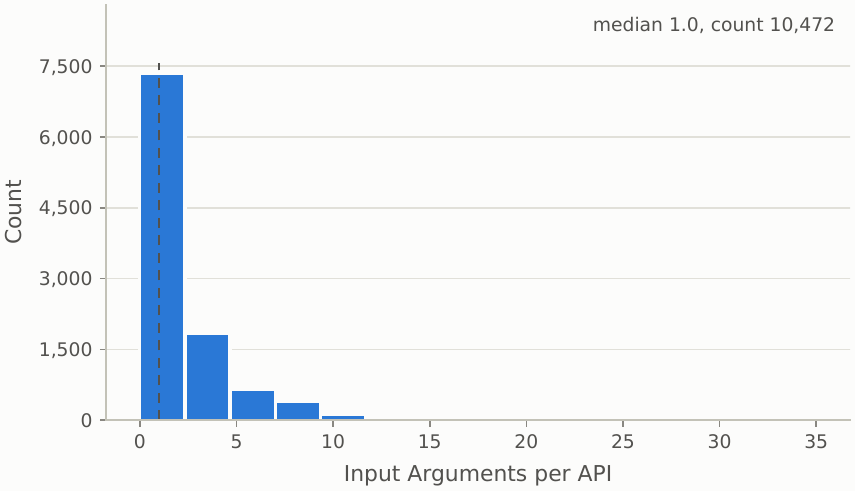}
    \caption{Input Arguments / API}
    \label{fig-app-info-args}
  \end{subfigure}
  \hfill
  \begin{subfigure}[b]{0.32\textwidth}
    \includegraphics[width=\textwidth]{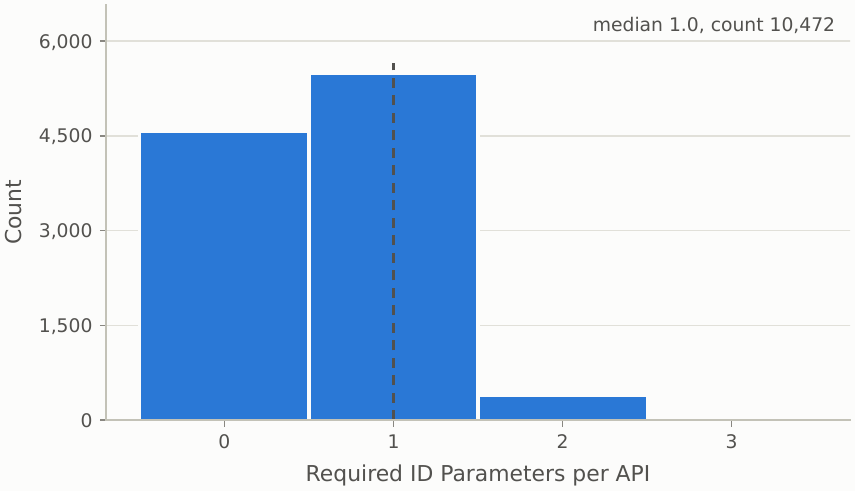}
    \caption{Required IDs / API}
    \label{fig-app-info-ids}
  \end{subfigure}
  \hfill
  \begin{subfigure}[b]{0.32\textwidth}
    \includegraphics[width=\textwidth]{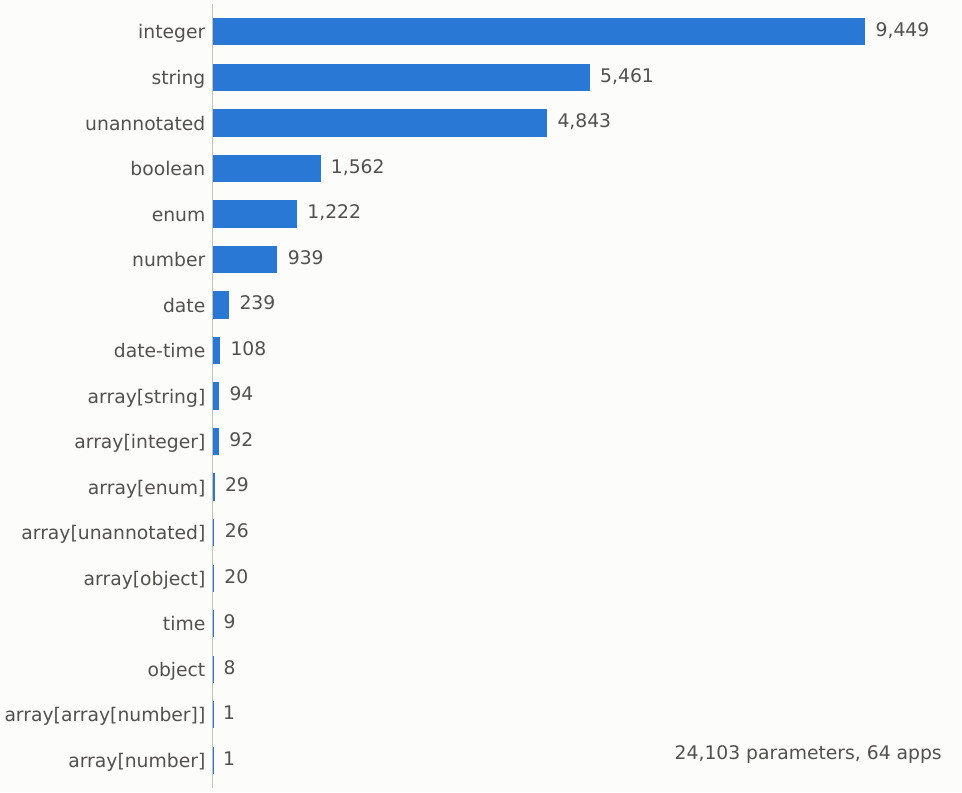}
    \caption{Parameter Types}
    \label{fig-app-info-types}
  \end{subfigure}
  \caption{
      Statistics of the generated apps and their APIs. The top row shows per-app statistics and the bottom row shows per-API and per-parameter statistics. The sample used here is the complete set of 64 apps used in all the experiments.
  }
  \label{fig-app-info}
\end{figure}

In this section, we detail the statistics on the overall sets of apps and tasks generated by \method.
First, in \Cref{fig-app-info} we visualize statistics of the 64 generated apps and the APIs they expose.
\Cref{fig-app-info-apis} shows that apps expose a median of 144 APIs, totaling 10,472 APIs across all apps, of which 6,266 are write and 4,206 are read APIs.
\Cref{fig-app-info-classes} shows that apps use a median of 55 distinct data classes, with some apps using over 120.
\Cref{fig-app-info-funcs} shows that apps implement a median of 81 distinct functionalities.
\Cref{fig-app-info-args} shows that most APIs take between 1 and 10 input arguments.
\Cref{fig-app-info-ids} shows that most APIs require at most one ID parameter referencing an existing database entity, with a small number requiring two or three.
This indicates that most APIs require composition and first calling other APIs to get the appropriate ID data.
\Cref{fig-app-info-types} shows the distribution of types across all 24,103 API parameters showing there is a diversity of data types beyond strings.
Unannotated types refer to dict data types.

\section{Training Details}
\label{supp-training-details}

For each task set $(\mathcal{A}_i, \mathcal{T}_i)$ we roll out GLM-5.2 \citep{glm5team2026glm5} on every task through the same interface and prompt as the agent to collect data.
We use vLLM \citep{kwon2023efficient} for sampling and use the default sampling parameters.
The GLM expert attempts each task up to 4 times and we save the first trajectory where the task verifier returns True indicating the task was successfully completed.
Therefore, each task is only represented with 1 training trajectory in the SFT dataset.
When collecting the data we save the reasoning trace from the data collection policy as well for supervision.

\begin{figure}[t!]
    \centering
    \begin{subfigure}[b]{0.48\linewidth}
        \centering
        \includegraphics[width=\linewidth]{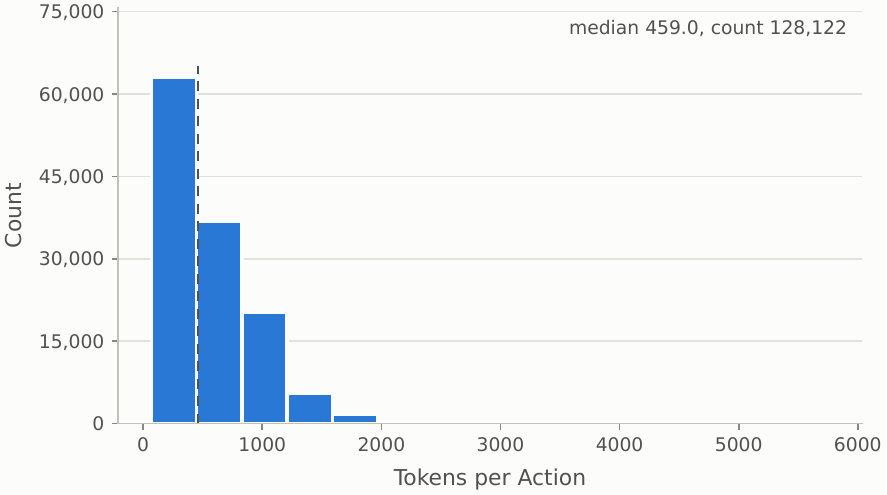}
        \caption{Tokens per action}
        \label{fig:tokens-per-action}
    \end{subfigure}
    \hfill
    \begin{subfigure}[b]{0.48\linewidth}
        \centering
        \includegraphics[width=\linewidth]{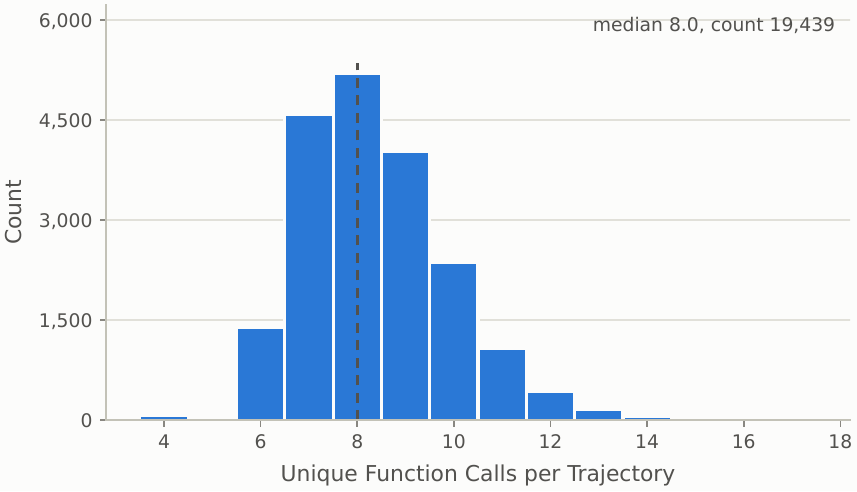}
        \caption{Unique function calls per trajectory}
        \label{fig:unique-fn-calls}
    \end{subfigure}
    \caption{
        Statistics of the GLM-5.2 expert trajectories in the SFT dataset, both computed on a subsample of the overall set of training trajectories.
        \textbf{(a)} Distribution of the number of tokens per expert action.
        \textbf{(b)} Distribution of the number of unique function calls per trajectory.
    }
    \label{fig:trajectory-stats}
\end{figure}

\Cref{fig:trajectory-stats} visualizes data about the resulting trajectories. 
The trajectory actions have a median of 459 tokens per action block.
The action which is generated after the reasoning includes some preamble text and then a code block that invokes function calls from the tool library.
The actions call a median of 8 tools from the app tool libraries per trajectory, but the count ranges from 4 to 18.

We train the student on the collected data with next-token cross-entropy and apply the loss only to the expert's reasoning text and code blocks.
The agent prompt, $x$, tool results, and all other environment tokens are masked out.
We train with bf16, learning rate $2\times10^{-5}$, and $1\%$ learning rate warmup, max gradient norm of 1.0 and sequence packing. For the 2B model we use pack sequence length 262,144, for 4B we use 262,144 and 2 steps of gradient accumulation, for 9B we use 2 steps of gradient accumulation and pack size 131,072, and for 27B we use pack size 65,536 and gradient accumulation 4 and also train on $2\times$ more nodes than for the other sizes.
We train on 8xB200 nodes and use Flash Attention 4 \citep{zadouri2026flashattention4}.
Unless otherwise specified, we run every experiment on 1 node, and scale the \sapiFull experiments to $2\times$ the number of nodes.

We hold out a random subset of 4k trajectories, spanning different apps, to use as validation data. 
We evaluate the model on the validation data every 200 updates.
The experiments in \Cref{sec-task-properties} and \Cref{sec-scaling} train for at most 1 epoch over the dataset.
However, if the validation loss increases 5 times in a row across successive validation runs, we early stop the training run before the target number of epochs.
The \sapiFull experiment in \Cref{sec-baseline} increases to 4 epochs and uses $2\times$ the amount of compute by scaling the number of nodes but keeps the other settings the same.

\section{Analysis Details}
\label{supp-analysis-details}

In all scaling experiments, we fit \cref{eq:appgen-scaling-law} independently per model size, minimizing the unweighted sum of squared residuals between predicted and observed aggregate score with the trust-region reflective solver in SciPy.
We hold out the runs in the top 25\% of training FLOPs and fit on the remaining runs, so the held-out set lies outside the compute range used for fitting and tests extrapolation.
\Cref{tab:scaling-law-coefficients} shows the fitted parameters.

\begin{table}[ht]
\centering
\caption{Fitted scaling-law coefficients of \cref{eq:appgen-scaling-law} by model size.}
\label{tab:scaling-law-coefficients}
\begin{tabular}{c|rrrrr}
\hline
\textbf{Model size}
& $\mathbf{c_0}$
& $\mathbf{c_A}$
& $\mathbf{b_0}$
& $\mathbf{b_T}$
& $\boldsymbol{\nu}$ \\
\hline
2B  & $-2.46$ & 0.181 &  0.54   & 0.595 & 20.00 \\
4B  & $-0.57$ & 0.247 & $-0.66$ & 0.401 &  1.67 \\
9B  & $-0.25$ & 0.236 & $-0.41$ & 0.495 &  1.46 \\
27B &   0.20  & 0.392 & $-9.96$ & 1.475 &  0.05 \\
\hline
\end{tabular}
\end{table}

\section{Benchmark and Evaluation Details}
\label{supp-benchmark}
In each benchmark environment, we enable the agent to use the same agent interface of \Cref{app:agent-interface}.
Each benchmark's tools are exposed to the agent through the meta-tools of \Cref{app:agent-interface}, so \texttt{list\_apps}, \texttt{list\_tools}, and \texttt{get\_tool\_documentation} serve that benchmark's own API documentation.
The agent executes the tools by emitting Python code.
Any called environment tools are then dispatched to the environment.
We use the same agent prompt as used during training.

In AppWorld, we report the task goal completion (TGC) and restrict evaluation to 50 steps maximum.
In ASTRA-Bench we restrict the evaluation to only the hardest complexity of each of the three task categories.
In BFCL we use v3 and the multi-turn split only with the default benchmark settings.
For evaluation, we use the default Qwen3.5 sampling parameters in vLLM.
We restrict the maximum number of per-action output tokens to 4096 including the reasoning and the action.

\section{Additional Results}
\label{app:additional-results}

\subsection{Additional Analysis}
\label{supp-analysis}
\label{supp-alloc-analysis}

\newcommand{\allocpanel}[1]{%
    \IfFileExists{#1}{%
        \includegraphics[width=\linewidth]{#1}%
    }{%
        \fbox{\begin{minipage}[b][0.62\linewidth][c]{\dimexpr\linewidth-2\fboxsep-2\fboxrule\relax}%
            \centering\scriptsize\color{gray}not yet generated%
        \end{minipage}}%
    }%
}
\newcommand{\alloccolhead}[1]{%
    \begin{minipage}[b]{0.32\linewidth}\centering\small #1\end{minipage}%
}

We show all the per model size app-to-task allocation plots below.
We also report the fit quality of the scaling law in \cref{eq:appgen-scaling-law} for each student size, both on the runs used for fitting (in-sample) and on all held-out runs in the top 25\% of training FLOPs (extrapolation), in units of the normalized aggregate score.
For the 2B student, the in-sample RMSE is $0.043$ and MAE is $0.033$, and the extrapolation RMSE and MAE are both $0.004$.
For the 4B student, the in-sample RMSE is $0.050$ and MAE is $0.039$, and the extrapolation RMSE is $0.066$ and MAE is $0.051$.
For the 9B student, the in-sample RMSE is $0.050$ and MAE is $0.038$, and the extrapolation RMSE is $0.033$ and MAE is $0.026$.
For the 27B student, the in-sample RMSE is $0.076$ and MAE is $0.049$, and the extrapolation RMSE is $0.037$ and MAE is $0.028$.

\begin{figure}[h!]
    \centering
    \begin{subfigure}[b]{0.32\linewidth}
        \centering
        \includegraphics[width=\linewidth]{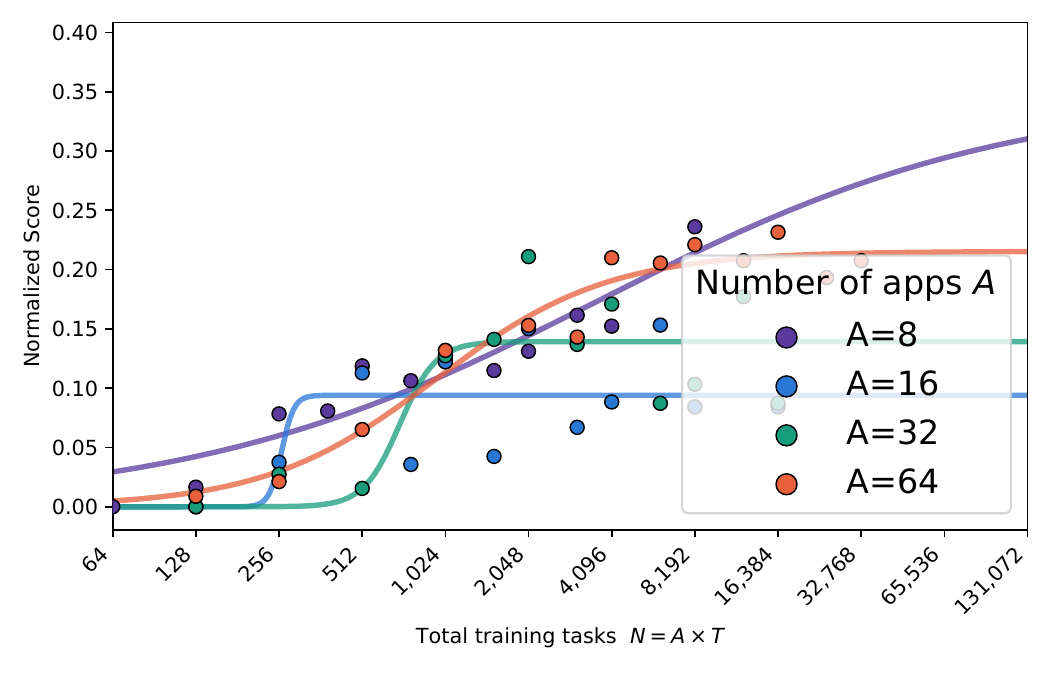}
        \caption{App vs.\ task allocation}
    \end{subfigure}
    \hfill
    \begin{subfigure}[b]{0.32\linewidth}
        \centering
        \includegraphics[width=\linewidth]{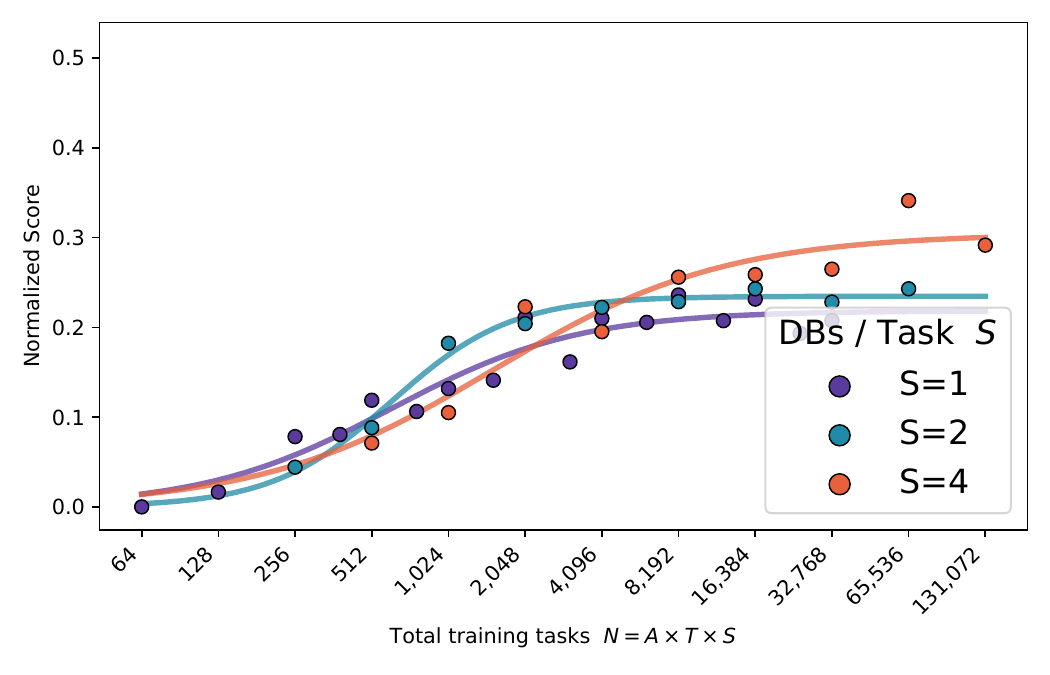}
        \caption{Database breadth}
    \end{subfigure}
    \hfill
    \begin{subfigure}[b]{0.32\linewidth}
        \centering
        \includegraphics[width=\linewidth]{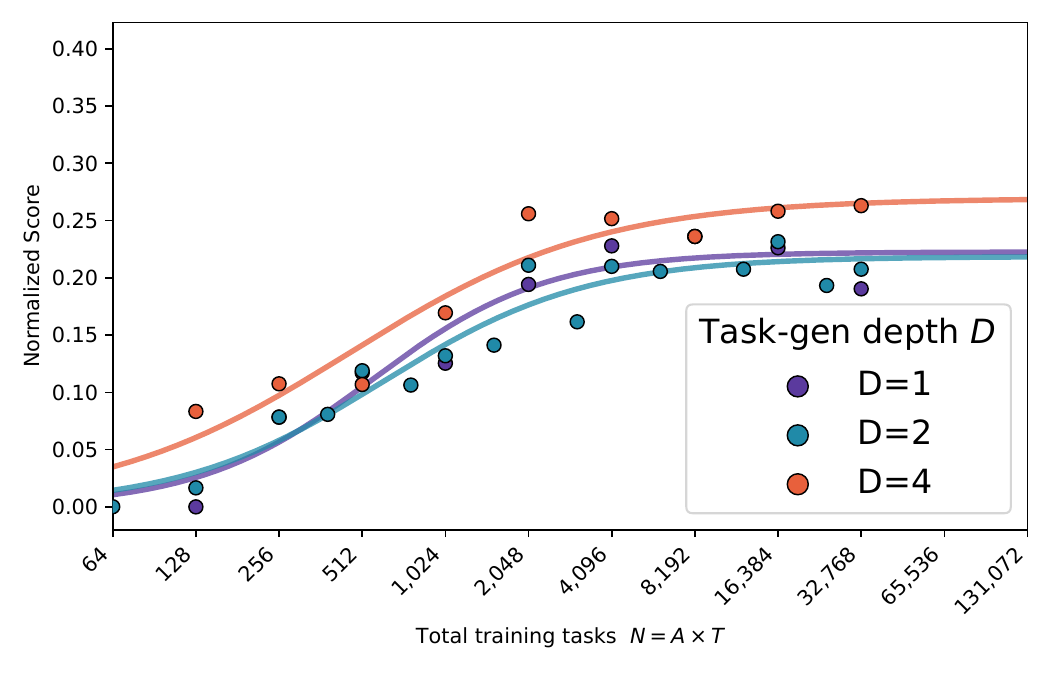}
        \caption{Database depth}
    \end{subfigure}
    \caption{
        Allocation experiment for the 2B model.
    }
    \label{fig-allocation-2b}
\end{figure}

\begin{figure}[h!]
    \centering
    \begin{subfigure}[b]{0.32\linewidth}
        \centering
        \includegraphics[width=\linewidth]{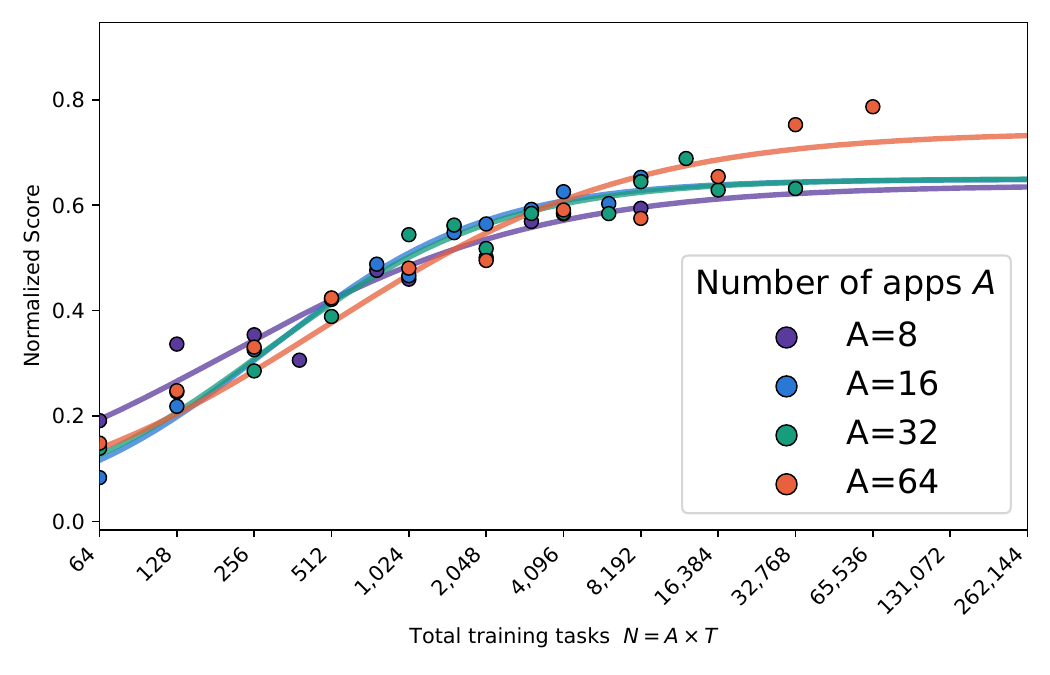}
        \caption{App vs.\ task allocation}
    \end{subfigure}
    \hfill
    \begin{subfigure}[b]{0.32\linewidth}
        \centering
        \includegraphics[width=\linewidth]{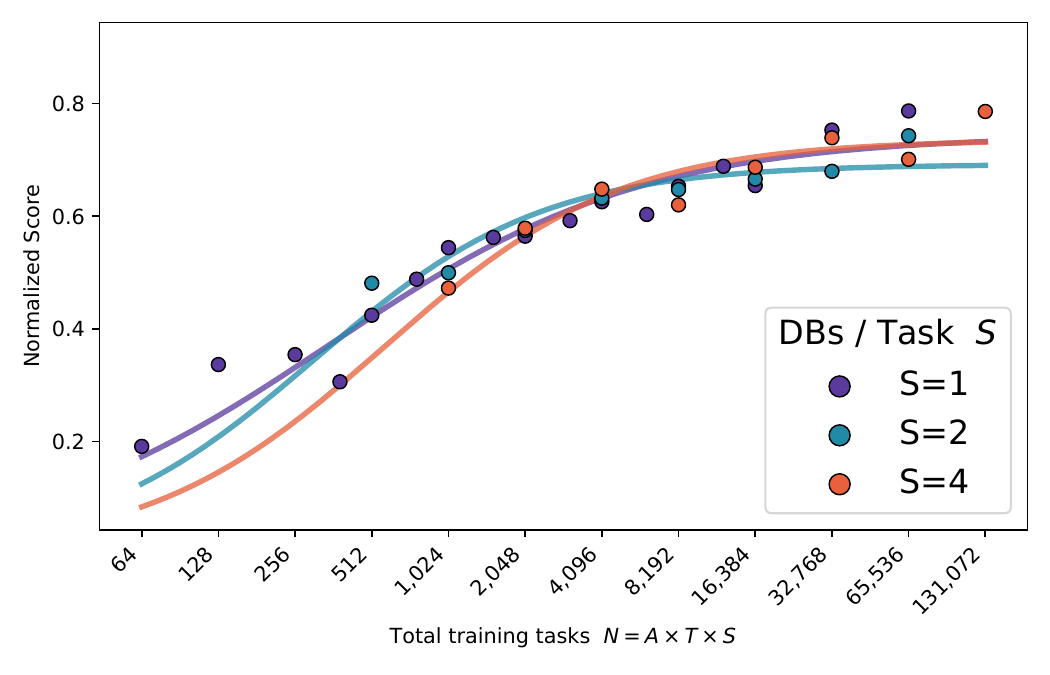}
        \caption{Database breadth}
    \end{subfigure}
    \hfill
    \begin{subfigure}[b]{0.32\linewidth}
        \centering
        \includegraphics[width=\linewidth]{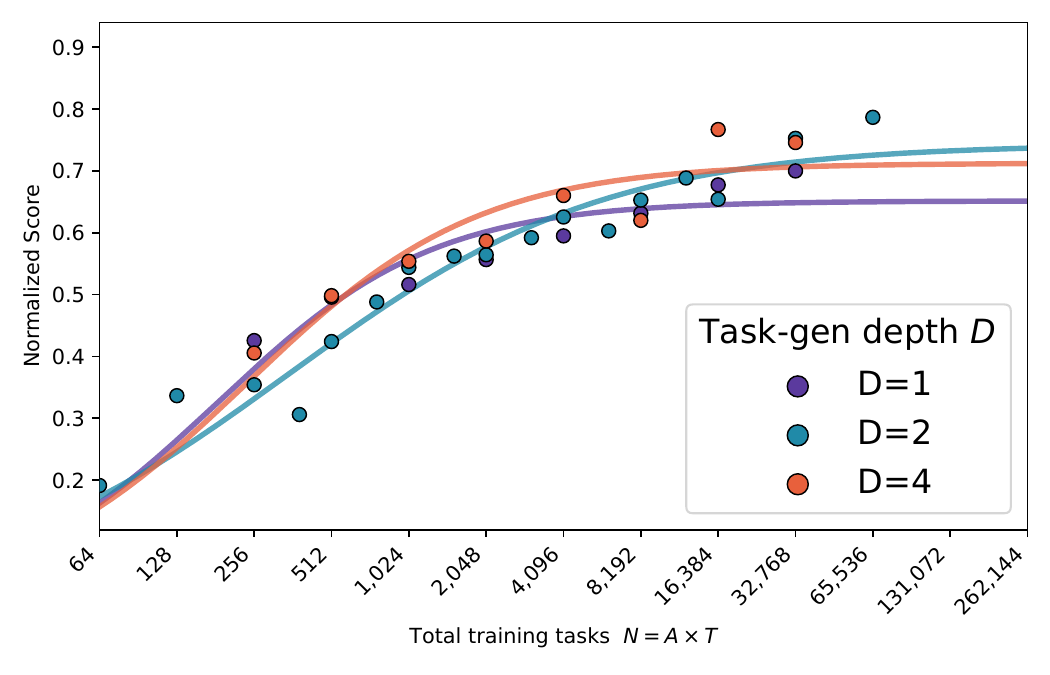}
        \caption{Database depth}
    \end{subfigure}
    \caption{
        Allocation experiment for the 4B model.
    }
    \label{fig-allocation-4b}
\end{figure}

\begin{figure}[h!]
    \centering
    \begin{subfigure}[b]{0.32\linewidth}
        \centering
        \includegraphics[width=\linewidth]{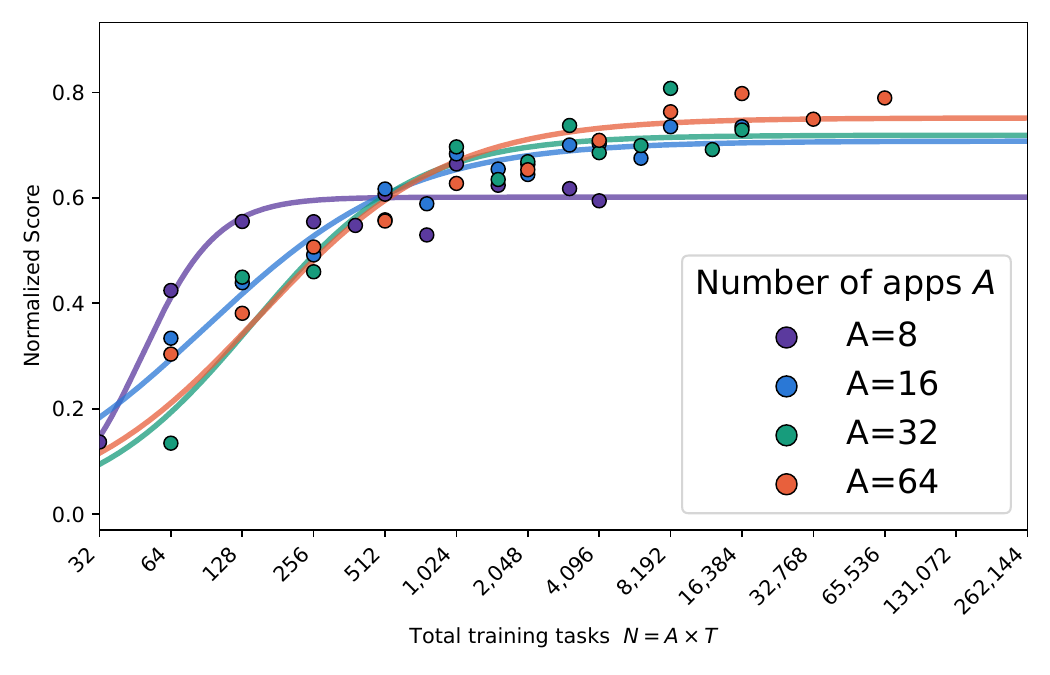}
        \caption{App vs.\ task allocation}
    \end{subfigure}
    \hfill
    \begin{subfigure}[b]{0.32\linewidth}
        \centering
        \includegraphics[width=\linewidth]{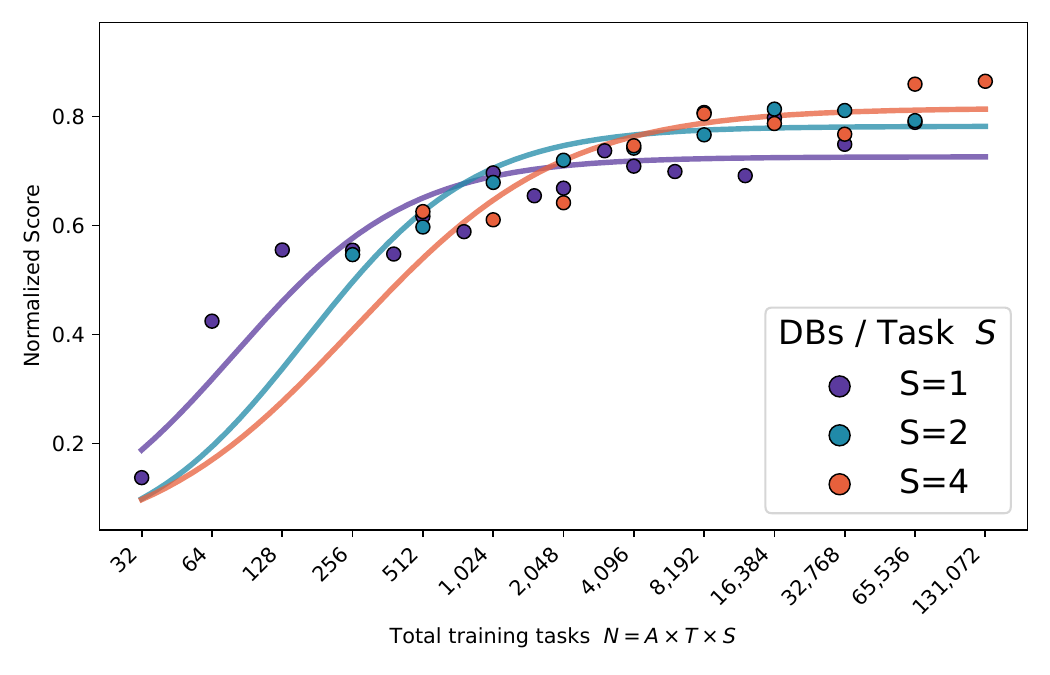}
        \caption{Database breadth}
    \end{subfigure}
    \hfill
    \begin{subfigure}[b]{0.32\linewidth}
        \centering
        \includegraphics[width=\linewidth]{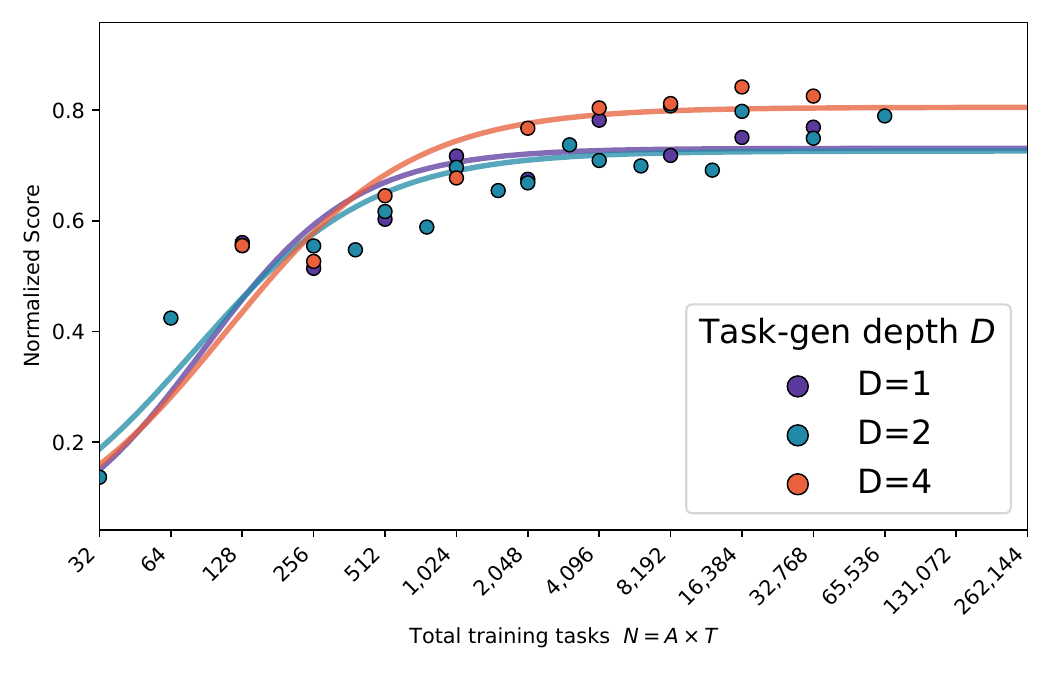}
        \caption{Database depth}
    \end{subfigure}
    \caption{
        Allocation experiment for the 9B model.
    }
    \label{fig-allocation-9b}
\end{figure}

\begin{figure}[h!]
    \centering
    \begin{subfigure}[b]{0.32\linewidth}
        \centering
        \includegraphics[width=\linewidth]{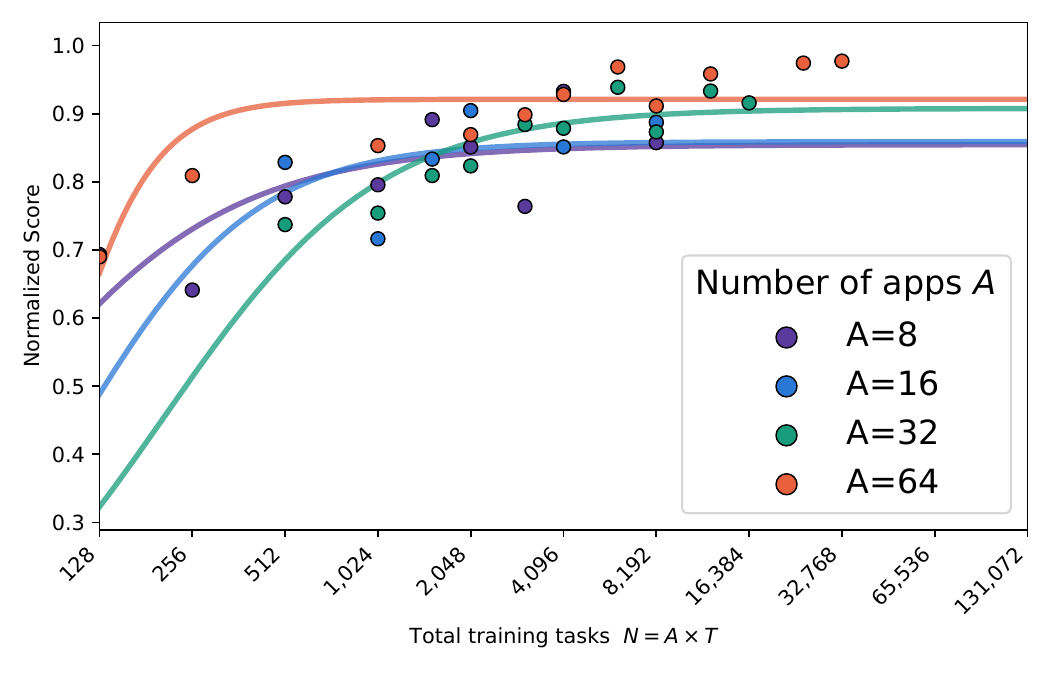}
        \caption{App vs.\ task allocation}
    \end{subfigure}
    \hfill
    \begin{subfigure}[b]{0.32\linewidth}
        \centering
        \includegraphics[width=\linewidth]{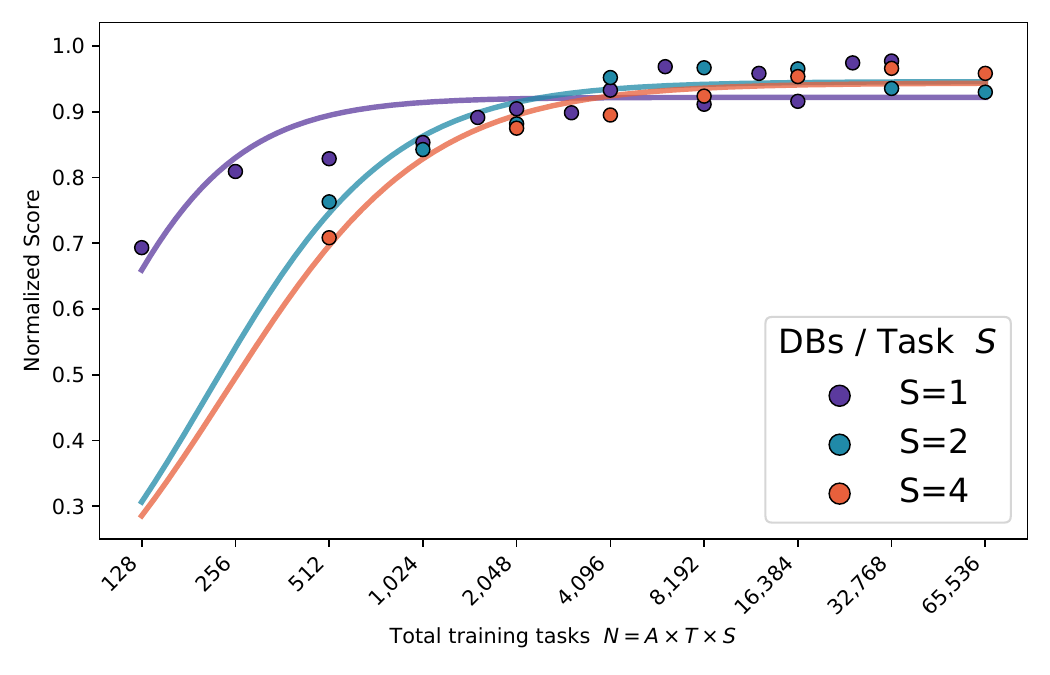}
        \caption{Database breadth}
    \end{subfigure}
    \hfill
    \begin{subfigure}[b]{0.32\linewidth}
        \centering
        \includegraphics[width=\linewidth]{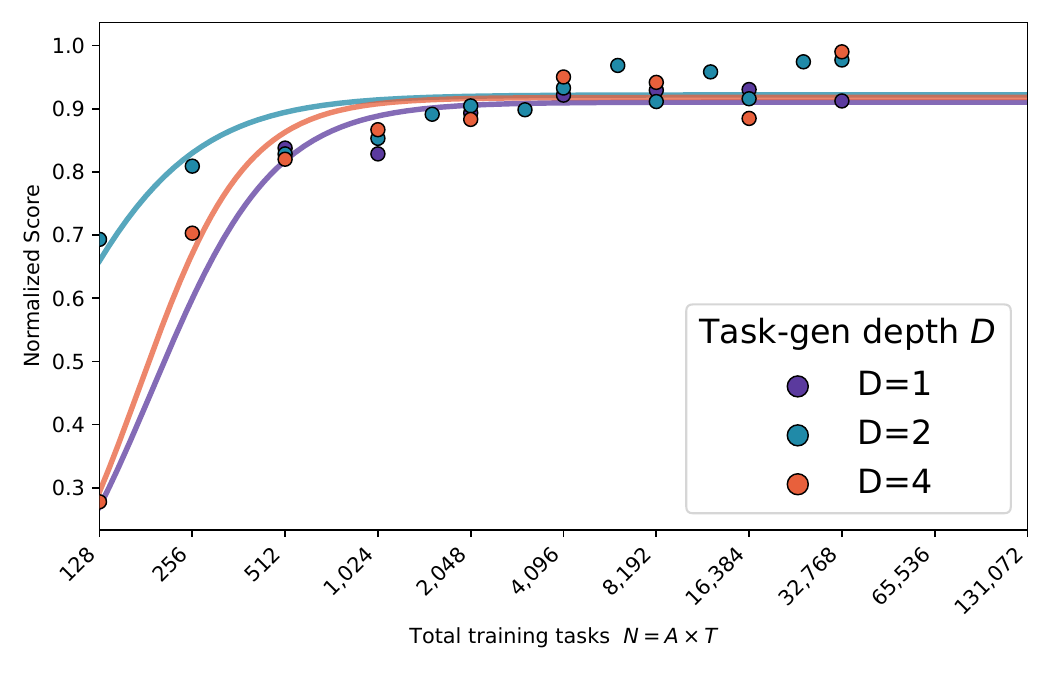}
        \caption{Database depth}
    \end{subfigure}
    \caption{
        Allocation experiment for the 27B model.
    }
    \label{fig-allocation-27b}
\end{figure}

\subsection{Benchmark Results Breakdown}
\label{supp-results-breakdown}

\Cref{tab:baseline-comparison} reports a single aggregate number per benchmark and model size.
\Cref{tab:appworld-breakdown,tab:astrabench-breakdown,tab:sapi-holdout-breakdown} give the full breakdown behind those aggregates: both AppWorld splits, and the per-category scores on ASTRA-Bench and the \method holdout.

\begin{table}[h]
    \centering
    \begin{tabular}{lcccccc}
        \toprule
        & \multicolumn{2}{c}{Qwen3.5-4B} & \multicolumn{2}{c}{Qwen3.5-9B} & \multicolumn{2}{c}{Qwen3.5-27B} \\
        \cmidrule(lr){2-3} \cmidrule(lr){4-5} \cmidrule(lr){6-7}
        Approach & Test-N & Test-C & Test-N & Test-C & Test-N & Test-C \\
        \midrule
        Zero-shot      & 47.5 & 14.7 & 45.4 & 26.4 & 81.8 & 61.5 \\
        SimSAT         & 66.8 & 58.8 & 74.5 & 65.4 & -  & -    \\
        Nemotron       & 26.0 & 17.5 & 31.9 & 20.8 & -  & -    \\
        ToolACE        & 20.2 & 10.9 & 18.6 & 12.4 & -  & -    \\
        ToolAlpaca     & 25.7 & 17.3 & 24.4 & 15.7 & -  & -    \\
        \midrule
        \sapiSmallSub  & 39.3 & 38.4 & 56.5 & 53.0 & 78.6 & 73.8 \\
        \sapiSmallOpt  & 47.0 & 45.0 & 58.2 & 56.2 & 81.0 & 74.1 \\
        \sapiFull      & \textbf{81.0} & \textbf{66.4} & \textbf{83.3} & \textbf{77.0} & \textbf{90.6} & \textbf{81.5} \\
        \bottomrule
    \end{tabular}
    \caption{
        Full AppWorld results on the Test-Normal (Test-N) and Test-Challenge (Test-C) splits.
        The SimSAT, ToolACE, and ToolAlpaca results are from \citet{lee2026environment}.
    }
    \label{tab:appworld-breakdown}
\end{table}

\begin{table}[h]
    \centering
    \begin{tabular}{llcccc}
        \toprule
        Model & Approach & Ref. & Func. & Info. & Overall \\
        \midrule
        \multirow{5}{*}{Qwen3.5-4B}
          & Zero-shot     & 55.5 & 33.5 & 60.3 & 57.4 \\
          & Nemotron      & 50.1 & 27.5 & 58.6 & 54.4 \\
          & \sapiSmallSub & 69.7 & 48.1 & 74.4 & 71.1 \\
          & \sapiSmallOpt & 71.4 & 51.0 & 74.8 & 72.3 \\
          & \sapiFull     & \textbf{73.7} & \textbf{53.2} & \textbf{76.3} & \textbf{74.2} \\
        \midrule
        \multirow{5}{*}{Qwen3.5-9B}
          & Zero-shot     & 69.5 & 50.3 & 74.2 & 71.8 \\
          & Nemotron      & 64.6 & 47.2 & 70.4 & 67.7 \\
          & \sapiSmallSub & 70.4 & 49.1 & 78.2 & 74.2 \\
          & \sapiSmallOpt & 71.9 & 52.1 & 78.9 & 75.5 \\
          & \sapiFull     & \textbf{75.5} & \textbf{53.2} & \textbf{81.1} & \textbf{76.8} \\
        \midrule
        \multirow{5}{*}{Qwen3.5-27B}
          & Zero-shot     & 79.1 & 62.4 & 82.7 & 80.2 \\
          & Nemotron      & -    & -    & -    & -    \\
          & \sapiSmallSub & 74.1 & 55.7 & 80.4 & 77.3 \\
          & \sapiSmallOpt & 78.6 & 58.8 & 81.3 & 79.3 \\
          & \sapiFull     & \textbf{82.8} & \textbf{64.7} & \textbf{85.4} & \textbf{82.3} \\
        \bottomrule
    \end{tabular}
    \caption{Full ASTRA-Bench results broken down by task category. Note that in ASTRA-Bench, a single task can count for multiple categories, thus the overall score is not the average of the individual columns.}
    \label{tab:astrabench-breakdown}
\end{table}

\begin{table}[h]
    \centering
    \small
    \begin{tabular}{llccccc}
        \toprule
        Model & Approach & Shop. & Msg. & Fin. & Music & Avg. \\
        \midrule
        \multirow{5}{*}{Qwen3.5-4B}
          & Zero-shot     & 0.0 & 0.0 & 1.0 & 0.0 & 0.3 \\
          & Nemotron      & 0.0 & 0.0 & 1.0 & 3.0 & 1.0 \\
          & \sapiSmallSub & 50.0 & 51.0 & \textbf{60.0} & 44.0 & 51.3 \\
          & \sapiSmallOpt & \textbf{64.0} & \textbf{74.0} & 36.0 & \textbf{50.0} & \textbf{56.0} \\
          & \sapiFull     & 60.0 & 72.0 & 32.0 & 48.0 & 53.0 \\
        \midrule
        \multirow{5}{*}{Qwen3.5-9B}
          & Zero-shot     & 1.0 & 1.0 & 3.0 & 1.0 & 1.5 \\
          & Nemotron      & 0.0  & 1.0  & 0.0  & 1.0  & 0.5  \\
          & \sapiSmallSub & 64.0 & 72.0 & 40.0 & 48.0 & 56.0 \\
          & \sapiSmallOpt & 71.0 & \textbf{79.0} & 47.0 & \textbf{55.0} & 63.0 \\
          & \sapiFull     & \textbf{76.0} & 76.0 & \textbf{52.0} & 52.0 & \textbf{64.0} \\
        \midrule
        \multirow{5}{*}{Qwen3.5-27B}
          & Zero-shot     & 8.0 & 35.0 & 15.0 & 22.0 & 20.0 \\
          & Nemotron      & -    & -    & -    & -    & -    \\
          & \sapiSmallSub & 72.0 & 76.0 & 56.0 & 52.0 & 64.0 \\
          & \sapiSmallOpt & 74.0 & \textbf{78.0} & 64.0 & \textbf{56.0} & 68.0 \\
          & \sapiFull     & \textbf{80.0} & 77.0 & \textbf{68.0} & 54.0 & \textbf{69.8} \\
        \bottomrule
    \end{tabular}
    \caption{Full \method-Holdout results broken down by app category.}
    \label{tab:sapi-holdout-breakdown}
\end{table}

\subsection{Results on Another Data Synthesis Pipeline}
\label{supp-simsat}

\begin{wrapfigure}{r}{0.38\textwidth}
    \centering
    \includegraphics[width=\linewidth]{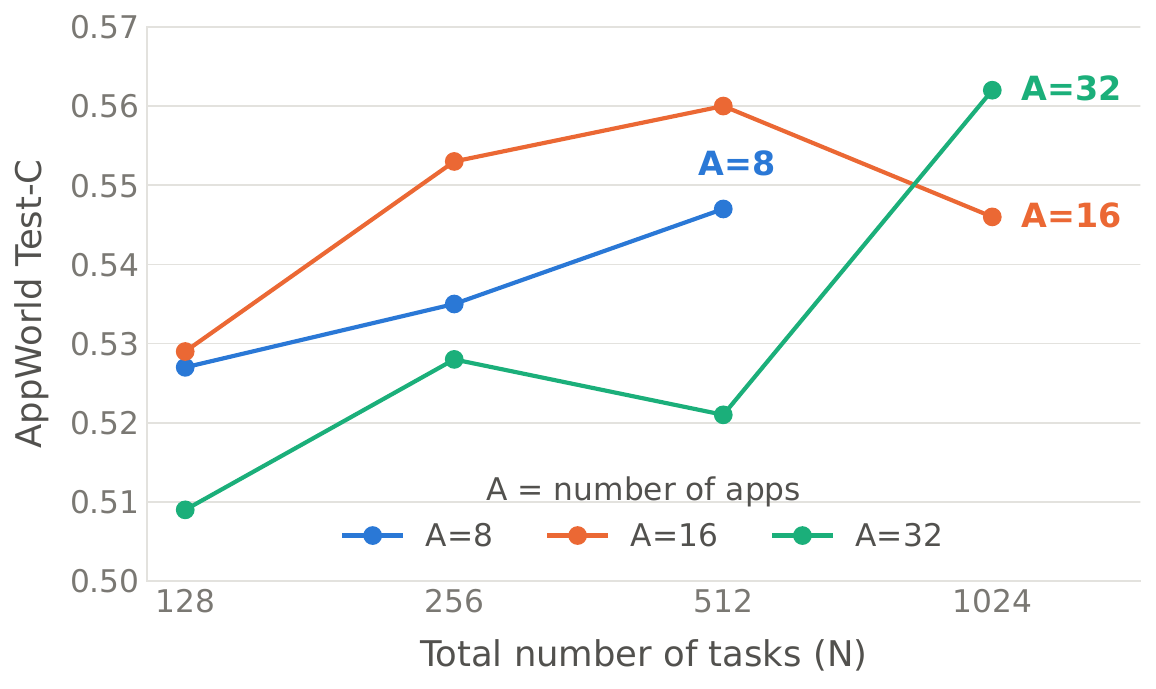}
    \caption{
        Comparing the allocation between apps and tasks per-app on another data synthesis pipeline SimSAT~\citep{lee2026environment}.
    }
    \label{fig:simsat}
\end{wrapfigure}

Here, we show that the findings we observe in \Cref{sec-task-properties}---on how tool diversity (measured by number of apps) affects the scaling behavior as we increase the total number of tasks---hold not only on our \method pipeline, but are also valid on another data synthesis pipeline. Specifically, we validate our findings with another synthetic tool-use dataset, SimSAT~\citep{lee2026environment}.
SimSAT contains synthetic tasks and their corresponding tool-use trajectories generated from 50 different apps. Under fixed task budgets of $N = \{128, 256, 512, 1024\}$, we sample various sets of tasks generated from varying number of apps ($A = \{8, 16, 32\}$), and study how downstream performance changes. As shown in \Cref{fig:simsat}, we see that under lower task budgets with $N = \{128, 256, 512\}$, it is favorable to generate more tasks per app ($A=32$ generally underperforms compared to $A = \{8, 16\}$). However, with a higher task budget at $N = 1024$, we see an opposite trend that generating tasks from more apps unlocks better performance, comparing $A=32$ to $A=16$. This aligns with our observation on \method, where high tool diversity matters as we scale the number of training tasks.

\end{document}